\RequirePackage[OT1]{fontenc}
\documentclass[letterpaper, 10 pt, conference]{ieeeconf}
\IEEEoverridecommandlockouts
\usepackage{ifthen}
\newcommand\arxivmode{true} 
\ifthenelse{\equal{\arxivmode}{true}}{\pdfoutput=1}{}

\usepackage[utf8]{inputenc}
\usepackage{pgfplots}
\usepgfplotslibrary{groupplots,dateplot}
\usetikzlibrary{patterns,shapes.arrows}
\pgfplotsset{compat=newest}

\usepackage{tikz}
\usetikzlibrary{external,positioning}
\usepgfplotslibrary{external}
\makeatletter
\let\NAT@parse\undefined
\makeatother
\usepackage[numbers]{natbib} 
\usepackage{subfig} 
\usepackage[ruled,linesnumbered]{algorithm2e} 
\usepackage{graphics} 
\usepackage{epsfig} 
\usepackage{amsmath} 
\usepackage{amsfonts} 
\usepackage[capitalise]{cleveref} 
\usepackage[per-mode=symbol]{siunitx}

\usepackage{amssymb}  
\usepackage{balance} 
\usepackage{verbatim} 
\usepackage{xcolor} 
\usepackage{float}

\newcommand{\edit}[1]{{\color{black}{#1}}}
\newcommand{\wennie}[1]{{\color{black}{#1}}}

\newcommand{\ogmap}{\mathbf{m}}
\newcommand{\numcells}{| \ogmap |}
\newcommand{\cell}{m}
\newcommand{\pocc}{o}
\newcommand{\broi}{b}
\newcommand{\dcell}{d}
\newcommand{\meas}{\mathbf{z}}
\newcommand{\hmeas}{\meas^h}
\newcommand{\rmeas}{\meas^r}
\newcommand{\rmeasc}{\hat{\meas}^r}
\newcommand{\ogmapc}{\hat{\ogmap}}
\newcommand{\roicells}{\mathbf{c}}
\newcommand{\roitraversedcells}{\hat{\roicells}}
\newcommand{\numcellsc}{| \ogmapc |}
\newcommand{\cellc}{\hat{\cell}}
\newcommand{\chit}{\cell_{\text{hit}}}
\newcommand{\csqmi}{I_{\text{CS}}}
\newcommand{\oavi}{I_{\text{OAVI}}}
\newcommand{\uaware}{I_{\text{UA}}}
\newcommand{\iroi}{I_{\text{ROI}}}
\newcommand{\aroi}{\alpha_{\text{ROI}}}
\newcommand{\proxaware}{I_{\text{PA}}}
\newcommand{\apa}{\alpha_{\text{PA}}}
\newcommand{\vizlikeli}{P_{\text{V}}} 

\title{\LARGE \bf
  Collaborative Human-Robot Exploration via Implicit Coordination
}

\author{Yves Georgy Daoud, Kshitij Goel, Nathan Michael, and Wennie Tabib
\thanks{The authors are with the Robotics Institute, Carnegie Mellon University, Pittsburgh, PA 15213 USA
(e-mail: \{ydaoud,kshitij,nmichael,wtabib\}@cmu.edu).}%
}

\begin{document}

\maketitle
\thispagestyle{empty}
\pagestyle{empty}

\begin{abstract}
  This paper develops a methodology for collaborative human-robot
  exploration that leverages implicit coordination. Most autonomous single- and
  multi-robot exploration systems require a remote operator to provide
  explicit guidance to the robotic team. Few works consider how \edit{to embed the
    human partner alongside} robots to provide guidance in the field.
  A remaining challenge for collaborative human-robot exploration is efficient
  communication of goals from the human to the robot. In this paper we
  develop a methodology that implicitly communicates a region of interest from
  a helmet-mounted depth camera on the human's head to the robot and
  an information gain-based exploration objective that biases motion
  planning within the viewpoint provided by the human. The result is an aerial system
  that safely accesses regions of interest that may not be immediately
  viewable or reachable by the human. The approach is evaluated in simulation
  and with hardware experiments in a motion capture arena. Videos of the
  simulation and hardware experiments are available at: https://youtu.be/7jgkBpVFIoE.
\end{abstract}

\section{INTRODUCTION}
\begin{figure}
  \centering
  \subfloat[\label{sfig:1a}]{\includegraphics[width=\linewidth]{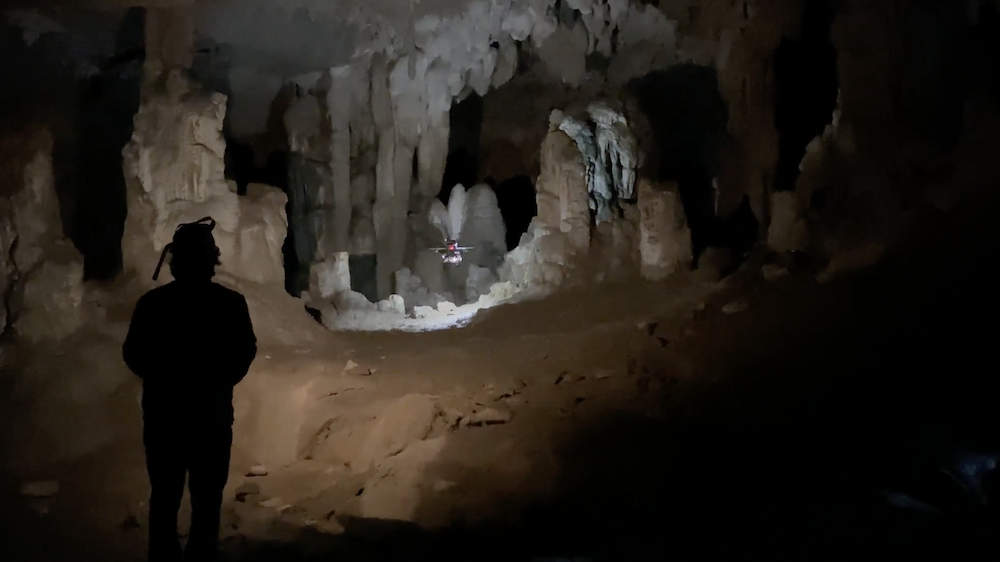}}\\
  \subfloat[\label{sfig:1b}]{\includegraphics[width=0.48\linewidth]{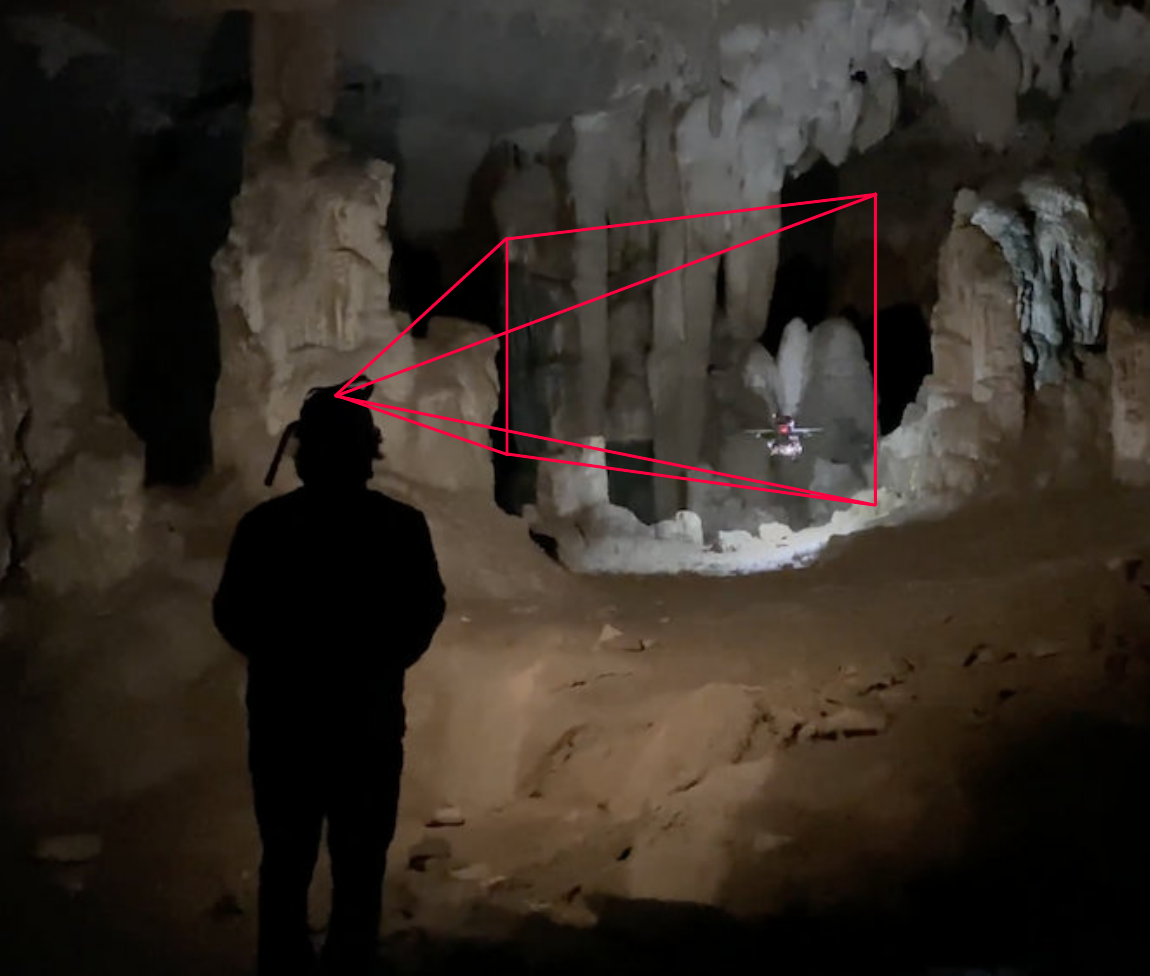}}\hfill
  \subfloat[\label{sfig:1c}]{\includegraphics[width=0.48\linewidth]{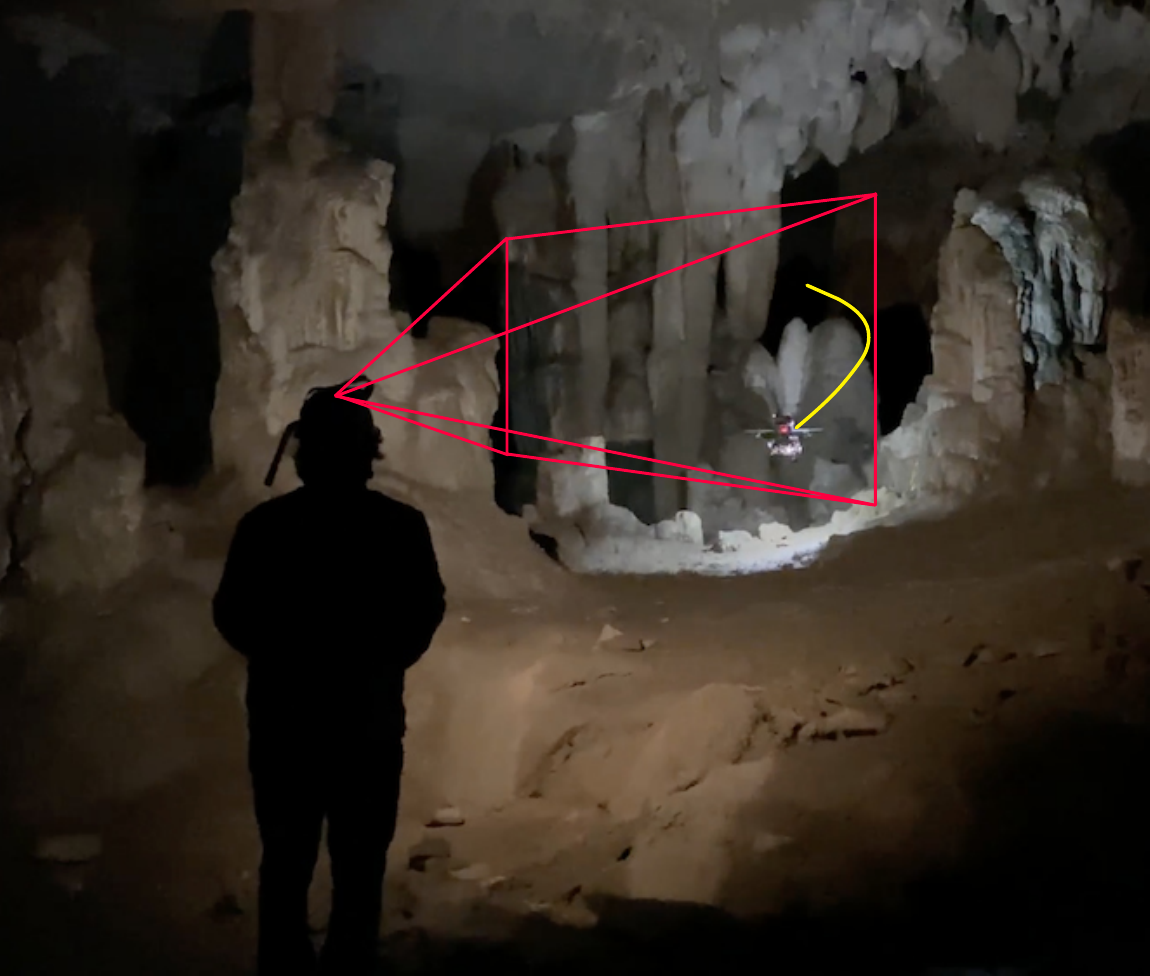}}
  \caption{\label{fig:1}
    \protect\subref{sfig:1a} A human-robot team is tasked with exploring a cave.
    \protect\subref{sfig:1b} The human implicitly conveys a region of interest to the robot by
    transmitting their current viewpoint.
  \protect\subref{sfig:1c} The robot plans a path to areas of the environment that are occluded to the human.}
\end{figure}

State-of-the-art exploration methodologies leverage the human as an
\emph{operator} outside of the exploration environment instead of directly
engaging them side-by-side with robots
\citep{cerberus,scherer2022resilient,tabib2019RSS}. Modeling the human as a
\emph{collaborator} instead of an \emph{operator} in a shared workspace for
exploration enables more efficient distributed exploration and useful emergent
robot behaviors~\citep{govindarajan2016human}.
In this work, a collaborative human-robot exploration system is developed
to explore 3D unstructured environments (\cref{sfig:1a}) by
communicating the field of view (FoV) of the human
to the robot (\cref{sfig:1b}) and having the robot use this
region of interest (ROI) to bias motion plans to
acquire views of \edit{areas} occluded to the human (\cref{sfig:1c}).

Explicit robot tasking is impractical~\cite{dominguez2021user} during
time-sensitive human-robot collaborative exploration (e.g. cave search
and rescue) if humans must reduce their operational
tempo~\citep{gu2022ar,keshavdas2013functional,wilde2020improving}, so
implicit communication of spatial goals is
imperative. State-of-the-art exploration objectives
reduce environmental uncertainty without providing the flexibility
to prioritize ROIs. To address these gaps in the state of the art,
this work presents a collaborative human-robot exploration
system that: (1) leverages implicit communication to spatially task an
aerial system to regions occluded to the human, and (2) develops an
information-gain based objective function inspired by the active object
reconstruction literature~\cite{delmerico2018informationcomparison} to
bias motion planning within the ROI specified by the human.
The approach is evaluated with real-time simulations and real-world hardware
experiments in a motion capture arena.

\section{RELATED WORK}
This work lies at the intersection of two key areas: implicit coordination for
collaborative human-robot exploration and motion planning objectives for active
mapping. In this section we review and contrast related works with
the method detailed in this paper.

Few prior works study implicit coordination for collaborative human-robot
exploration.~\citet{govindarajan2016human} achieve coordination through
a distributed strategy that assigns robots to homotopy classes
that are complementary to the ones being traversed by the human. It is assumed
that a blueprint of the environment is available to identify homotopy classes
before operation. In contrast, the proposed approach does not assume prior
information on the environment layout.
A motion primitive-based planner is leveraged to maximize
information gain, which takes the human's view into account, and
  drives the robot to explore regions occluded to the human; therefore,
  prior environment knowledge is not required.

  Within the context of multitasking, implicit communication has been
  used to augment human situational awareness via a robotic system.
  ~\citet{bentz2019unsupervised} leverage head tracking while a human
  performs an arbitrary number of complex tasks and fit the data to a visual
  interest function. An aerial robot uses
  the visual interest function to provide camera views that augment the human's
  situational awareness. This methodology does not directly translate
  to the exploration context because the visual interest function, which
  effectively rates the utility of a viewpoint, is difficult to specify before or during exploration.
  Instead, the proposed approach uses the notion of potential information gain over a discrete set of candidate
viewpoints to drive the robot towards the ROI.

\citet{reardon2019communicating} leverage augmented reality to share
information between a robot and human cooperatively exploring in the field.
The goal is to influence the behavior of the human teammate in the
human-robot cooperative exploration task by sharing information
about the robot's current plan, the task state, and communicating future actions.
In contrast, the proposed approach develops a methodology to influence the
robot's behavior depending on actions taken by the human. This implicit
coordination is desirable
in applications like search and rescue where the robot is expected to
adapt to the human's operational tempo.

Many motion planning objectives have been proposed for active mapping.
Frontier-based objectives utilize the distance to the boundary between unknown
and known space to drive the robot's exploration~\citep{yamauchi1998frontier}. For multi-agent operation, prioritization
between frontiers is utilized to assign agents towards complementary regions of
the environment~\citep{burgard2005coordinated}. However, deployments of this
idea and its variants have been limited to multiple
robots~\citep{dharmadhikari2021autonomous,scherer2022resilient}, with the human
largely supplying spatial goals explicitly when
desired~\citep{nevatia2008augmented}. Further, information-theoretic objectives utilize
the expected change in the entropy of the map due to candidate sensor
measurements to drive viewpoint selection in both
2D~\citep{bourgault2002information,julian2014mutual} and 3D
environments~\citep{charrow2015information}. While real-world operation has been
shown using single~\citep{tabib2021tro} and multiple aerial
robots~\citep{goel2021rapid}, these techniques have not been leveraged in
collaborative human-robot exploration. A key capability missing in these objectives
is the ability to prioritize spatial ROIs. Towards imposing such spatial
constraints, volumetric next-best-view (NBV) selection methods have been
proposed for the active object reconstruction problem where the viewpoint
is generated to focus on high-fidelity reconstruction of a single
object~\citep{vasquez2014volumetric,kriegel2015efficient}.
\citet{delmerico2018informationcomparison} propose several variants of
information gain objectives that are either
counting-based~\citep{vasquez2014volumetric},
probabilistic~\citep{kriegel2015efficient}, or a combination.
However, a method to apply these objectives in the collaborative human-robot
exploration system is lacking in the literature. To this end, this work
proposes and evaluates an Occlusion-Aware Volumetric Information (OAVI) objective
that extends the work of~\citet{delmerico2018informationcomparison} to the
collaborative human-robot exploration problem. We further contrast it to
ROI-constrained Cauchy-Schwarz Quadratic Mutual Information (ROI-CSQMI), an extension
of~\cite{charrow2015information} developed in this work, which applies \edit{the
human's FoV as a spatial constraint}.


\section{METHODOLOGY}
This section details the collaborative human-robot exploration method.
The human and robot incrementally build a shared map of the environment using range measurements, while
the robot uses the occupancy, ROI, and distance information within the shared map for
motion planning. We first describe the shared map representation.

\subsection{Shared Map Representation}\label{ssec:shared-map}
The shared map is modeled as a global 3D occupancy grid (OG) map, $\ogmap =\{
\cell_1, ..., \cell_{\numcells} \}$.  Each cell $\cell_i$ contains a tuple of
three scalar features: (1) the probability of occupancy ($\pocc_i$), (2)
a boolean
indicating if the cell is in \edit{the} ROI ($\broi_i$), and (3) the distance of the cell
from the closest obstacle ($\dcell_i$). Each cell $\cell_i$ is initially presumed
unknown ($\pocc_i = 0.5$), considered outside the ROI ($\broi_i = 0$), and
assumed to be at an infinite distance from the closest obstacle ($\dcell_i \rightarrow
\infty$). The range measurements at time $t$ are denoted by $\hmeas_t$ for the
human and $\rmeas_t$ for the robot. It is assumed that the global position and
orientation of these sensors are perfectly known.

\begin{figure}[h]
    \centering
    \subfloat[Human's FoV \label{sfig:humaninita}]{\includegraphics[width=0.23\textwidth]{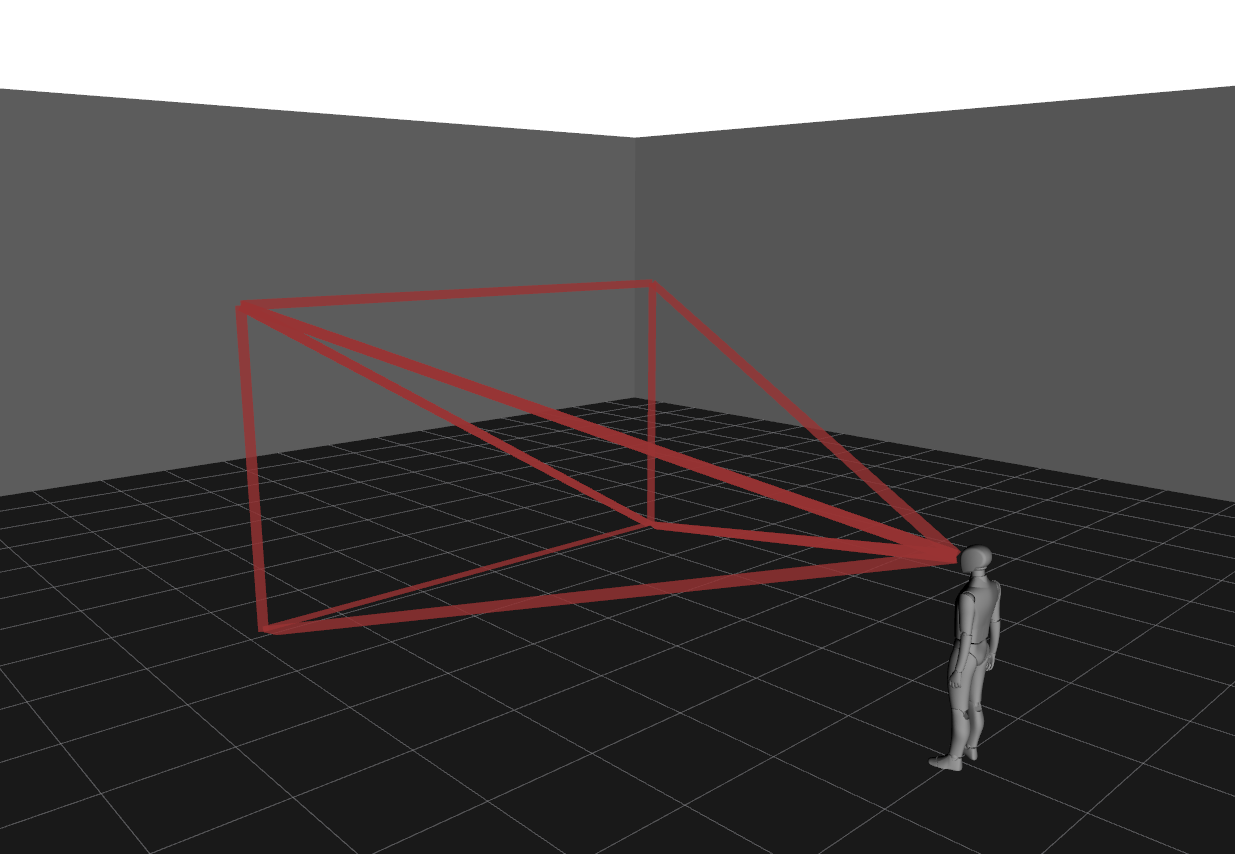}} \
    \subfloat[Voxels within the FoV \label{sfig:humaninitb}]{\includegraphics[width=0.23\textwidth]{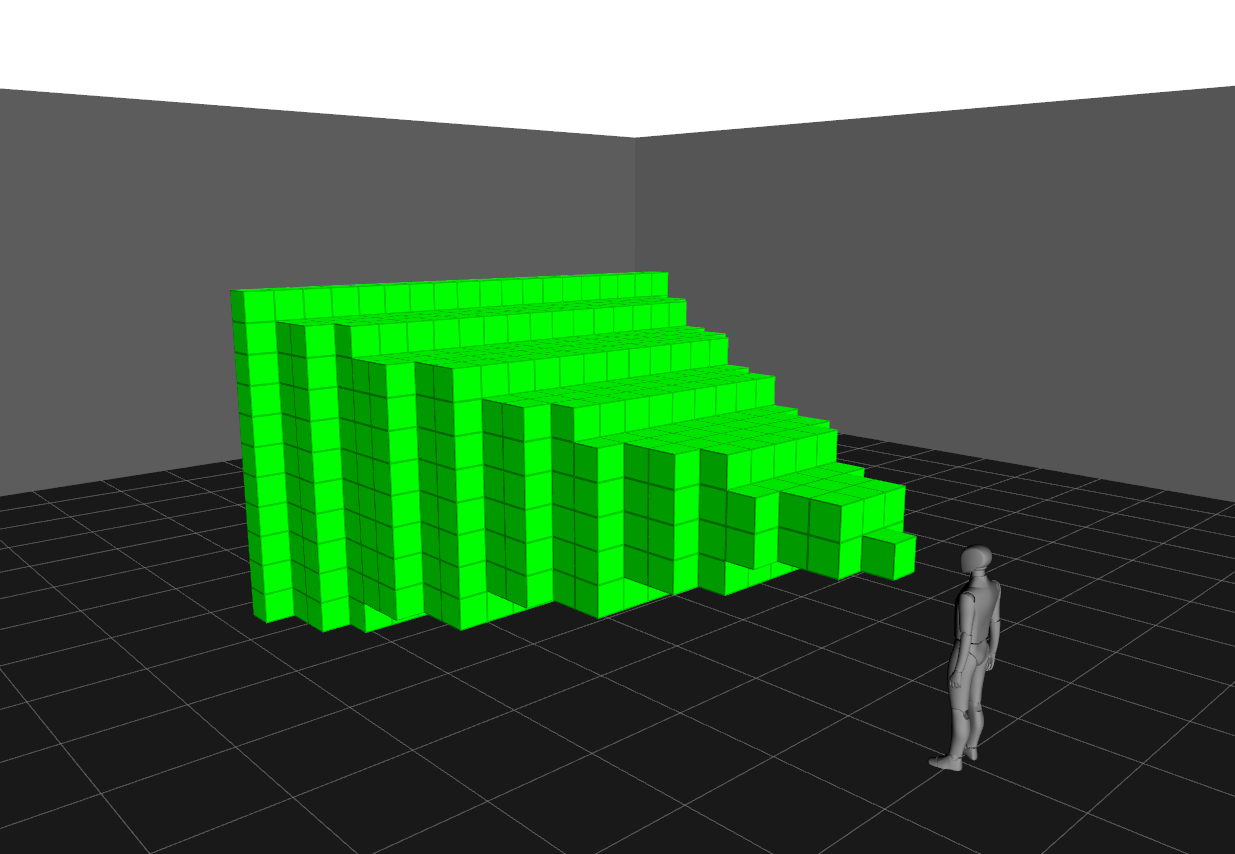}}
    \caption{\label{fig:humaninit}~\protect\subref{sfig:humaninita}
      The human's field of view (FoV) is \edit{shown in red}
      and used \edit{to determine which~\protect\subref{sfig:humaninitb}
      cells in the global occupancy map are within the ROI (shown in
      green)}.}
\end{figure}

For the cells within the FoVs of $\hmeas_t$ and $\rmeas_t$, the
probability of occupancy $\pocc_i$ is updated using the standard log-odds
update~\citep{thrun2002probabilistic}. However, the ROI values $\broi_i$ are
only set to $1$ within the FoV of $\hmeas_t$. The FoV is mathematically modeled
using the fusion of two triangles in 2D and two tetrahedrons in 3D built from
the sensor's intrinsic matrix (see~\cref{sfig:humaninita}). A subset
of cells corresponding to inliers of the FoV are shown in~\cref{sfig:humaninitb}.
This subset is extracted via inlier queries with
respect to the tetrahedrons on the centers of all cells in the shared map. The
distance values $\dcell_i$ are updated for the cells raycasted by both
$\hmeas_t$ and $\rmeas_t$ with the Euclidean distance from the nearest occupied
cell. We utilize the approximation by~\citet{delmerico2018informationcomparison}
that extends the rays behind a hit cell, $\chit$, and populates the distance
value at the current time, $\dcell^t_i$, for the remaining raycasted cells:
\begin{equation}
    \dcell^t_i =
    \begin{cases}
        \| f(\chit) - f(\cell_i) \|_2, & \text{if}\ \| f(\chit) - f(\cell_i) \|_2 < \dcell^{t-1}_i \\
        \dcell^{t-1}_i,         & \text{otherwise}
    \end{cases}
    \label{eq:dist}
\end{equation}
where $\dcell^{t-1}_i$ is the previously stored distance in cell $\cell_i$ and $f: \mathbb{Z}_{+} \rightarrow \mathbb{R}^n$ is a function that converts \wennie{a} cell index to the cell position in the world frame. $n = 2$ or $n = 3$ depending on the dimensionality of the map representation.

After obtaining the first observation from the human and updating the shared
map, the robot iteratively performs a two-step process: updating the map and
selecting the next best action. The rate of this process is specified by the user prior to
operation. The space of candidate actions used for action selection is generated
using a library of forward-arc motion primitives for a depth camera as presented
in \cite{tabib2021tro}. The best primitive
is chosen by maximizing the information gain over this discrete action space,
which is computed at the end viewpoint of each motion primitive.
We contribute \wennie{one information gain objective function (OAVI) and contrast to a baseline information-theoretic
objective function (CSQMI) as well as an extension of it (ROI-CSQMI) in this work}.

\begin{figure}
    \centering
    \subfloat[Environment\label{sfig:2Denva}]{\ifthenelse{\equal{\arxivmode}{true}}{\includegraphics[]{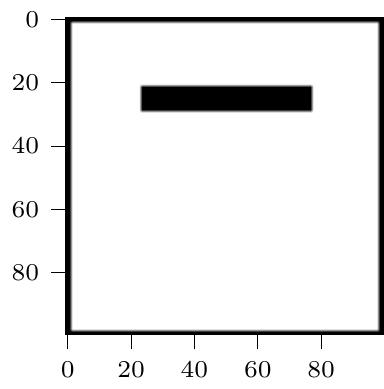}}{
\tikzsetnextfilename{2d_map}
\begin{tikzpicture}[font=\scriptsize]

\definecolor{darkgray176}{RGB}{176,176,176}

\begin{axis}[
width=0.27\textwidth,
height=0.27\textwidth,
tick align=outside,
tick pos=left,
x grid style={darkgray176},
xmin=-0.5, xmax=99.5,
xtick style={color=black},
y dir=reverse,
y grid style={darkgray176},
ymin=-0.5, ymax=99.5,
ytick style={color=black}
]
\addplot graphics [includegraphics cmd=\pgfimage,xmin=-0.5, xmax=99.5, ymin=99.5, ymax=-0.5] {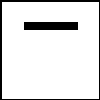};
\end{axis}

\end{tikzpicture}}} \
    \subfloat[Map after Human Observation\label{sfig:2Denvb}]{
\tikzsetnextfilename{2d_human_observation}
\begin{tikzpicture}[font=\scriptsize]

\definecolor{darkgray176}{RGB}{176,176,176}

\begin{axis}[
width=0.27\textwidth,
height=0.27\textwidth,
tick align=outside,
tick pos=left,
x grid style={darkgray176},
xmin=-0.5, xmax=99.5,
xtick style={color=black},
y dir=reverse,
y grid style={darkgray176},
ymin=-0.5, ymax=99.5,
ytick style={color=black}
]
\addplot graphics [includegraphics cmd=\pgfimage,xmin=-0.5, xmax=99.5, ymin=99.5, ymax=-0.5] {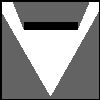};
\end{axis}

\end{tikzpicture}}\\
    \subfloat[CSQMI\label{sfig:csqmi}]{\ifthenelse{\equal{\arxivmode}{true}}{\includegraphics[]{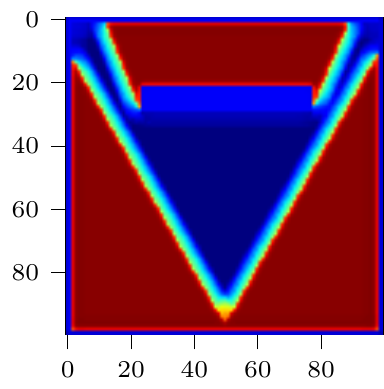}}{
\tikzsetnextfilename{2d_csqmi}
\begin{tikzpicture}[font=\scriptsize]

\definecolor{darkgray176}{RGB}{176,176,176}

\begin{axis}[
width=0.27\textwidth,
height=0.27\textwidth,
tick align=outside,
tick pos=left,
x grid style={darkgray176},
xmin=-0.5, xmax=99.5,
xtick style={color=black},
y dir=reverse,
y grid style={darkgray176},
ymin=-0.5, ymax=99.5,
ytick style={color=black}
]
\addplot graphics [includegraphics cmd=\pgfimage,xmin=-0.5, xmax=99.5, ymin=99.5, ymax=-0.5] {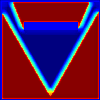};
\end{axis}

\end{tikzpicture}}} \
    \subfloat[ROI-CSQMI\label{sfig:roi-csqmi}]{\ifthenelse{\equal{\arxivmode}{true}}{\includegraphics[]{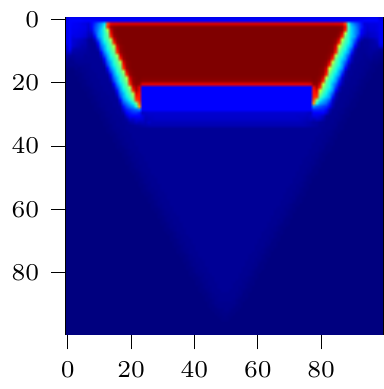}}{
\tikzsetnextfilename{2d_csqmi_with_roi}
\begin{tikzpicture}[font=\scriptsize]

\definecolor{darkgray176}{RGB}{176,176,176}

\begin{axis}[
width=0.27\textwidth,
height=0.27\textwidth,
tick align=outside,
tick pos=left,
x grid style={darkgray176},
xmin=-0.5, xmax=99.5,
xtick style={color=black},
y dir=reverse,
y grid style={darkgray176},
ymin=-0.5, ymax=99.5,
ytick style={color=black}
]
\addplot graphics [includegraphics cmd=\pgfimage,xmin=-0.5, xmax=99.5, ymin=99.5, ymax=-0.5] {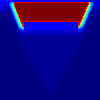};
\end{axis}

\end{tikzpicture}}}
    \caption{\label{fig:csqmi-numerical} Comparison of the
        information gain objectives using a 2D numerical example. For the environment
        in~\protect\subref{sfig:2Denva} and human at (50, 0), the map updated after one $\hmeas$ (\cref{ssec:shared-map}) is shown
        in~\protect\subref{sfig:2Denvb}. The CSQMI objective from~\citep{charrow2015information}
        shown in~\protect\subref{sfig:csqmi} does not account for $\hmeas$, while the
        \edit{ROI-CSQMI} objective in~\protect\subref{sfig:roi-csqmi} places higher weights
        in the occluded region.}
\end{figure}

\begin{figure*}
    \centering
    \subfloat[$\uaware$\label{sfig:oavi-uaware}]{\ifthenelse{\equal{\arxivmode}{true}}{\includegraphics[]{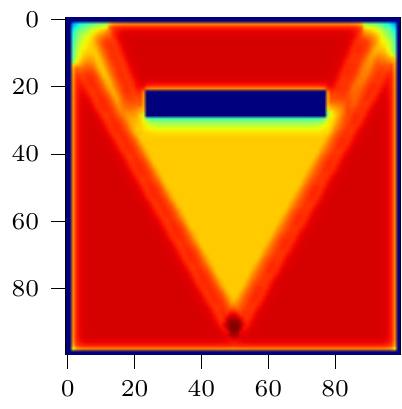}}{
\tikzsetnextfilename{2d_ioa}
\begin{tikzpicture}[font=\scriptsize]

\definecolor{darkgray176}{RGB}{176,176,176}

\begin{axis}[
width=0.28\textwidth,
height=5cm,
tick align=outside,
tick pos=left,
x grid style={darkgray176},
xmin=-0.5, xmax=99.5,
xtick style={color=black},
y dir=reverse,
y grid style={darkgray176},
ymin=-0.5, ymax=99.5,
ytick style={color=black}
]
\addplot graphics [includegraphics cmd=\pgfimage,xmin=-0.5, xmax=99.5, ymin=99.5, ymax=-0.5] {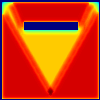};
\end{axis}

\end{tikzpicture}}} \
    \subfloat[$\iroi$\label{sfig:oavi-iroi}]{\ifthenelse{\equal{\arxivmode}{true}}{\includegraphics[]{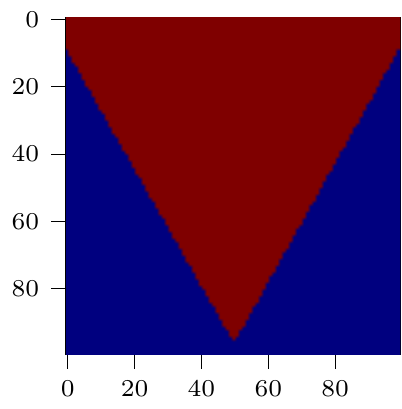}}{
\tikzsetnextfilename{2d_iroi}
\begin{tikzpicture}[font=\scriptsize]

\definecolor{darkgray176}{RGB}{176,176,176}

\begin{axis}[
width=0.28\textwidth,
height=5cm,
tick align=outside,
tick pos=left,
x grid style={darkgray176},
xmin=-0.5, xmax=99.5,
xtick style={color=black},
y dir=reverse,
y grid style={darkgray176},
ymin=-0.5, ymax=99.5,
ytick style={color=black}
]
\addplot graphics [includegraphics cmd=\pgfimage,xmin=-0.5, xmax=99.5, ymin=99.5, ymax=-0.5] {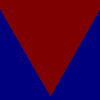};
\end{axis}

\end{tikzpicture}}} \
    \subfloat[$\proxaware$\label{sfig:oavi-proxaware}]{\ifthenelse{\equal{\arxivmode}{true}}{\includegraphics[]{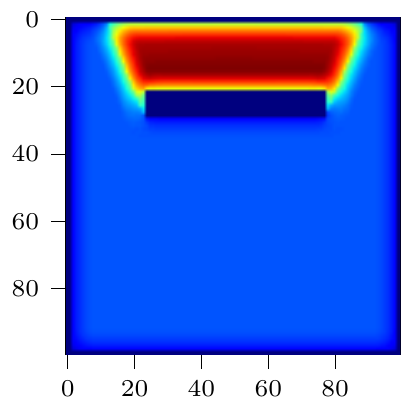}}{
\tikzsetnextfilename{2d_ipc}
\begin{tikzpicture}[font=\scriptsize]

\definecolor{darkgray176}{RGB}{176,176,176}

\begin{axis}[
width=0.28\textwidth,
height=5cm,
tick align=outside,
tick pos=left,
x grid style={darkgray176},
xmin=-0.5, xmax=99.5,
xtick style={color=black},
y dir=reverse,
y grid style={darkgray176},
ymin=-0.5, ymax=99.5,
ytick style={color=black}
]
\addplot graphics [includegraphics cmd=\pgfimage,xmin=-0.5, xmax=99.5, ymin=99.5, ymax=-0.5] {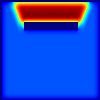};
\end{axis}

\end{tikzpicture}}} \
    \subfloat[OAVI\label{sfig:oavi}]{\ifthenelse{\equal{\arxivmode}{true}}{\includegraphics[]{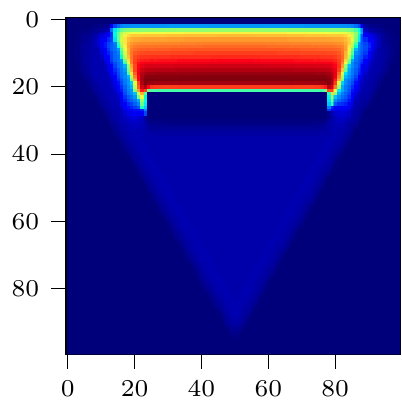}}{
\tikzsetnextfilename{2d_oae}
\begin{tikzpicture}[font=\scriptsize]

\definecolor{darkgray176}{RGB}{176,176,176}

\begin{axis}[
width=0.28\textwidth,
height=5cm,
tick align=outside,
tick pos=left,
x grid style={darkgray176},
xmin=-0.5, xmax=99.5,
xtick style={color=black},
y dir=reverse,
y grid style={darkgray176},
ymin=-0.5, ymax=99.5,
ytick style={color=black}
]
\addplot graphics [includegraphics cmd=\pgfimage,xmin=-0.5, xmax=99.5, ymin=99.5, ymax=-0.5] {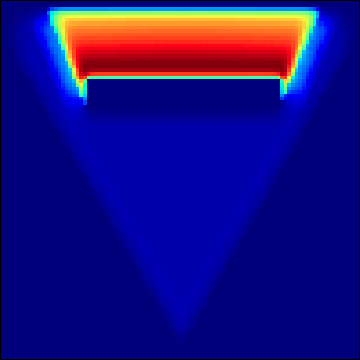};
\end{axis}

\end{tikzpicture}}}
    \caption{\label{fig:oavi-numerical} Heatmaps for the OAVI objective and its constituent
    terms (\cref{ssec:oavi}) over the 2D map shown in~\cref{sfig:2Denvb}.
    Compared to ROI-CSQMI in~\cref{sfig:roi-csqmi}, the OAVI objective function in~\cref{sfig:oavi}
    exhibits a gradient biasing the exploration to focus on the occluded region
    closer to the human's FoV first.}
\end{figure*}

\subsection{ROI-constrained CSQMI (ROI-CSQMI)}\label{ssec:roi-csqmi}
The ROI-CSQMI information gain objective \wennie{function modifies} the CSQMI objective function
proposed by~\citet{charrow2015information} \wennie{by imposing} a spatial
constraint corresponding to the FoV of the range measurement from the human
$\hmeas_t$. The original formulation proceeds as follows; \wennie{f}irst, for a candidate
viewpoint, a beam-based measurement model is used to determine which cells in
the current map will be observed via raycasting. Second, the CSQMI objective is
computed treating the raycasted cells independently of each other. Lastly, the CSQMI
contributions from all cells are added to determine the total utility of the viewpoint.
Our contribution lies in modifying the second step using the ROI information stored
in the shared map representation (\cref{ssec:shared-map}).

Let the set of cells raycasted by the candidate measurement $\rmeasc_t$ be
denoted by $\ogmapc$. This set is a subset of the current map, $\ogmapc
\subseteq \ogmap$. To impose the spatial constraint corresponding to the FoV
of $\hmeas_t$, we leverage the information in $\ogmap$ to obtain a
new set of raycasted cells within the ROI denoted as $\roitraversedcells$:
\begin{align}
    \roitraversedcells = \{ \cell_i \in \ogmapc \, | \, \broi_i = 1, i \in \{1, \ldots, \numcellsc\} \}.
    \label{eq:roi-csqmi-raycast}
\end{align}
\wennie{T}he ROI constraint \wennie{is imposed} at the
raycasting step \wennie{by finding the subset of cells that lie within
  the human's ROI}. The CSQMI objective for the candidate measurement $\rmeasc_t$,
$\csqmi[\ogmapc; \rmeasc_t]$, is computed using $\ogmapc$ via the equations
derived in~\citep{charrow2015information}. Note that when $\roitraversedcells = \ogmapc$,
the ROI-CSQMI objective function is equivalent to the CSQMI objective function.

For a 2D environment with an obstacle (\cref{sfig:2Denva}), the map
$\ogmap$ is shown in~\cref{sfig:2Denvb} after adding one range measurement
$\hmeas$ from the human \wennie{who is located} at the bottom center of the environment. \wennie{The grey,
black, and white cell colors denote unknown, occupied,
and free state, respectively}.~\Cref{sfig:csqmi} shows the heatmap for $\csqmi[\ogmapc; \rmeasc_t]$,
\wennie{which treats all unknown space equally}.~\Cref{sfig:roi-csqmi} shows the heatmap for $\csqmi[\roitraversedcells;
\rmeasc_t]$, \wennie{which prioritizes regions within the FoV of $\hmeas$}.
\wennie{The spatially constrained ROI-CSQMI objective enables implicit coordination between the human
and robot during} exploration by placing higher weight on views that intersect the ROI.
\wennie{However, there} are two
drawbacks \wennie{to} this modification: (1) the objective weighs all cells within
the occluded region equally, as opposed to the regions closer to the obstacle within the human's FoV, and (2)
once the robot enters the ROI during exploration, it is unlikely that it will \wennie{exit
it}. The OAVI objective, presented next, alleviates
these \wennie{drawbacks}.

\subsection{Occlusion-Aware Volumetric Information (OAVI)}\label{ssec:oavi}
\wennie{The proposed information-gain objective function, OAVI, is inspired by}
\cite{delmerico2018informationcomparison} \wennie{and modifies}
the uncertainty-aware\wennie{,} $\uaware$, the ROI\wennie{,} $\iroi$,
and the proximity-aware\wennie{,} $\proxaware$ \wennie{metrics}.

The uncertainty-aware metric $\uaware$
measures the uncertainty of the cell \wennie{and accounts} for potential
occlusions:
\begin{align}
    \uaware(\cell_i) =  H(\cell_i) \vizlikeli(\cell_i)\wennie{.}
    \label{eq:uaware}
\end{align}
$H(\cell_i)$ is Shannon's entropy~\citep{cover1999elements} of cell
$\cell_i$, and $\vizlikeli(\cell_i)$ is the likelihood \wennie{that the cell is visible}
from the current sensor pose. The result is shown in~\cref{sfig:oavi-uaware} for
the 2D map in~\cref{sfig:2Denvb} and \wennie{illustrates} high weights in the unknown space.

The ROI metric\wennie{,} $\iroi$\wennie{, biases the objective values} towards the ROI.
We \wennie{employ} the information stored in the shared map (\cref{ssec:shared-map}) to
mark the contribution of cells in the ROI towards $\iroi$ as $1$. For the other regions of the
map, the contribution is set to \wennie{a user-specified value,} $\aroi < 1$:
\begin{align}
    \iroi(\cell_i)=
    \begin{cases}
        1,     & \text{if } \broi_i = 1 \\
        \aroi, & \text{otherwise}
    \end{cases}
    \label{eq:iroi}
\end{align}
Intuitively, $\aroi$ controls the weight given to the unknown regions
of the environment outside the ROI. \wennie{A non-zero} $\aroi$ \wennie{enables the}
robot to \wennie{explore unknown regions} after prioritizing occluded regions within the ROI.
This metric is shown in~\cref{sfig:oavi-iroi} \wennie{with} $\aroi = 0.15$ for
the 2D map in~\cref{sfig:2Denvb}.

\wennie{The same modification is applied} for the proximity-aware metric\wennie{,} $\proxaware$\wennie{,
which} utilizes the distance values\wennie{,} $\dcell_i$\wennie{, in} the cells:
\begin{equation}
    \proxaware(\cell_i)=
    \begin{cases}
        d_{\max} - \dcell_i, & \text{if } \pocc_i = 0.5 \; \text{and} \; \dcell_i \leq d_{\max} \\
        \apa,     & \text{otherwise}
    \end{cases}
    \label{eq:I_PA}
\end{equation}
where $d_{max}$ is the max sensor range, $\dcell_i$ is the
distance from cell $\cell_i$ to the closest raytraced occupied cell, and $\apa \in [0, 1)$
is a tunable parameter. This modification produces a gradient
\wennie{that places higher weights on cells
close to an observed} surface \wennie{(e.g. see \cref{sfig:oavi-proxaware} where $\apa = 0.10$)}.

The final information gain $\oavi$ is defined as the cumulative product of \wennie{each metric}
over the raycasted cells $\ogmapc$ corresponding to the robot measurement at the viewpoint
$\rmeasc_t$:
\begin{align}
    \oavi[\ogmapc; \rmeasc_t] & = \sum_{i \in [1, \numcellsc]} \oavi(\cellc_i) \nonumber \\
       & = \sum_{i \in [1, \numcellsc]} \uaware(\cellc_i) \iroi(\cellc_i) \proxaware(\cellc_i).
    \label{eq:oavi}
\end{align}
\Cref{sfig:oavi} \wennie{illustrates} the heatmap corresponding to $\oavi[\ogmapc; \rmeasc_t]$.
\wennie{Note the gradient behind the obstacle in OAVI, which has the effect of weighting
  the viewpoints that observe occluded regions more heavily, and
  contrast this with the uniform weighting of ROI-CSQMI in~\cref{sfig:roi-csqmi}}.

\begin{figure*}
    \centering
    \subfloat[Single Wall Environment \label{sfig:simenvsa}]{\includegraphics[width=0.24\textwidth,trim=0 0 0 0,clip]{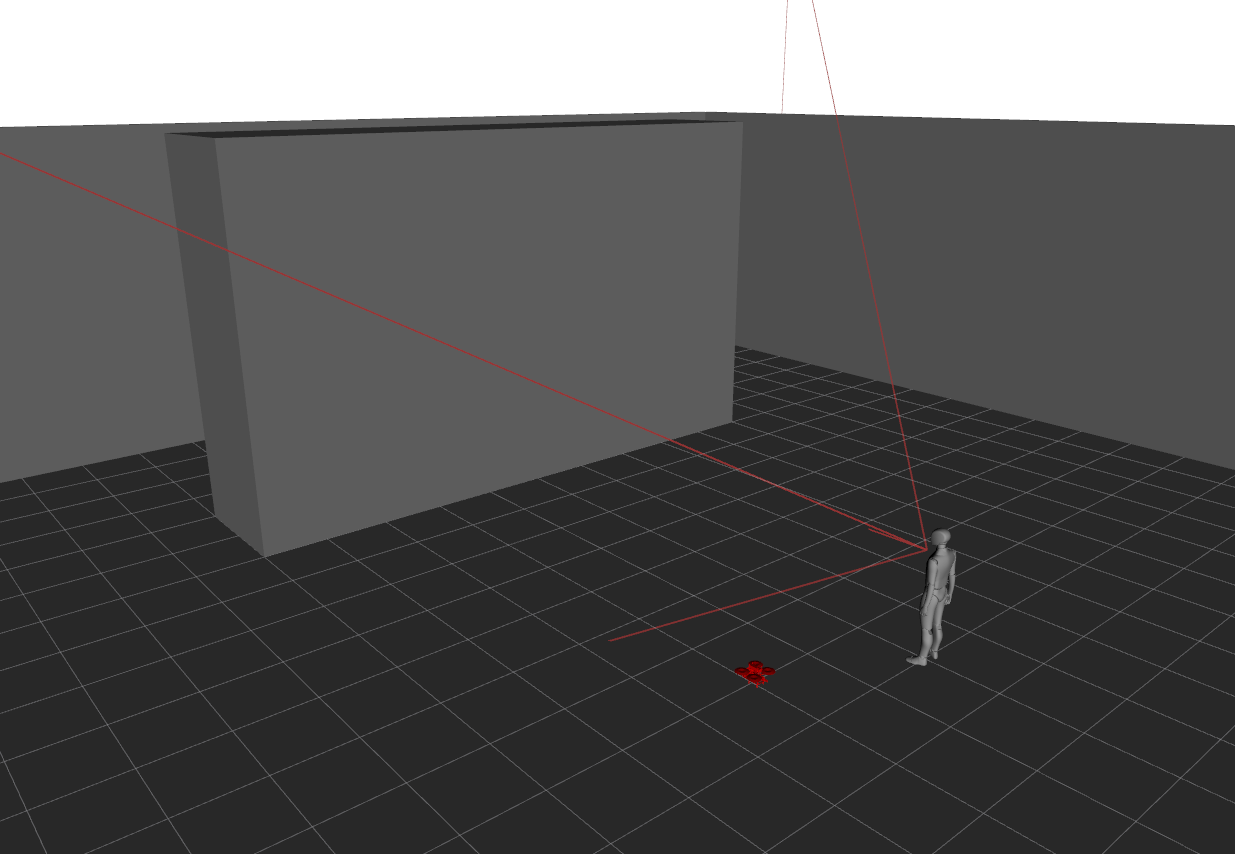}}\hfill
    \subfloat[Two Walls Environment \label{sfig:simenvsb}]{\includegraphics[width=0.24\textwidth,trim=0 0 0 0,clip]{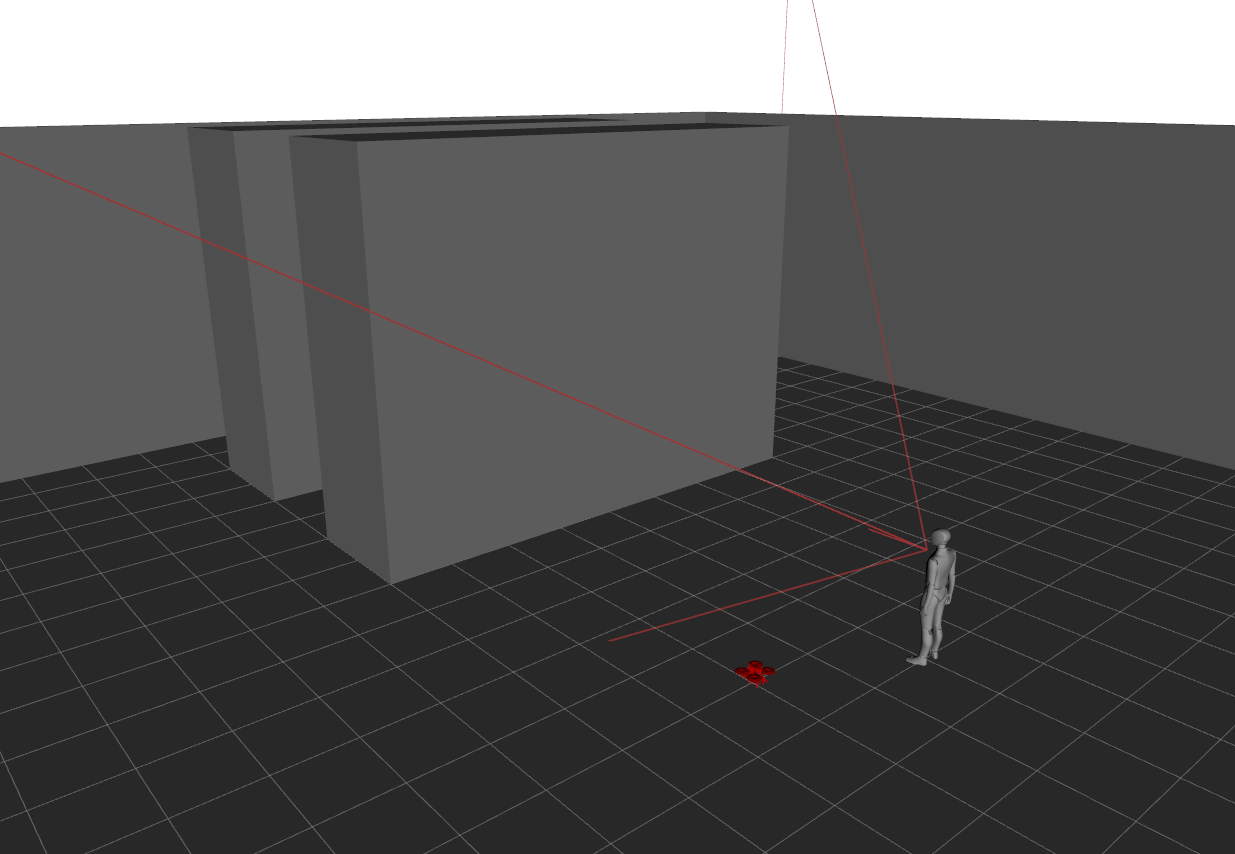}} \hfill
    \subfloat[Multiple Obstacles Environment \label{sfig:simenvsc}]{\includegraphics[width=0.24\textwidth,trim=0 0 0 0,clip]{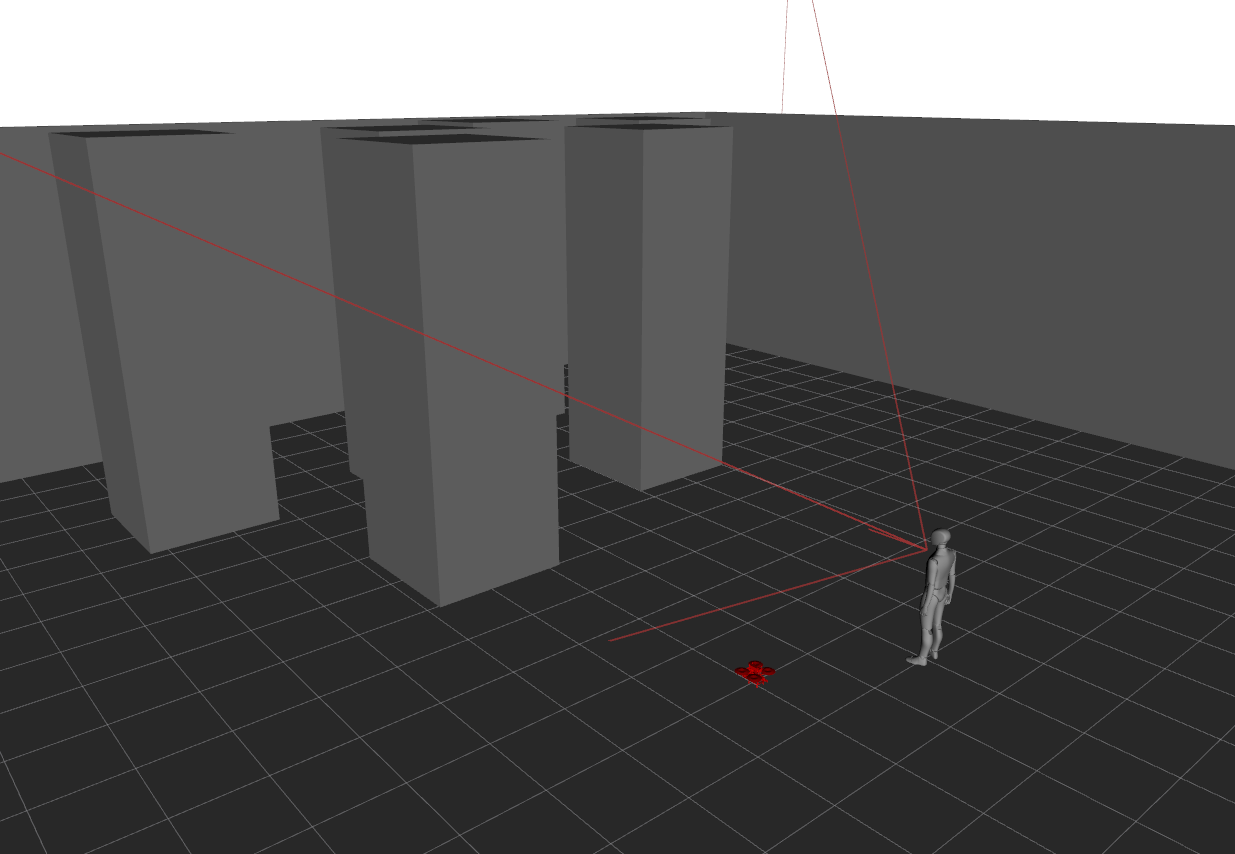}}\hfill
    \subfloat[Cave Environment \label{sfig:simenvsd}]{\includegraphics[width=0.24\textwidth,trim=0 0 0 0,clip]{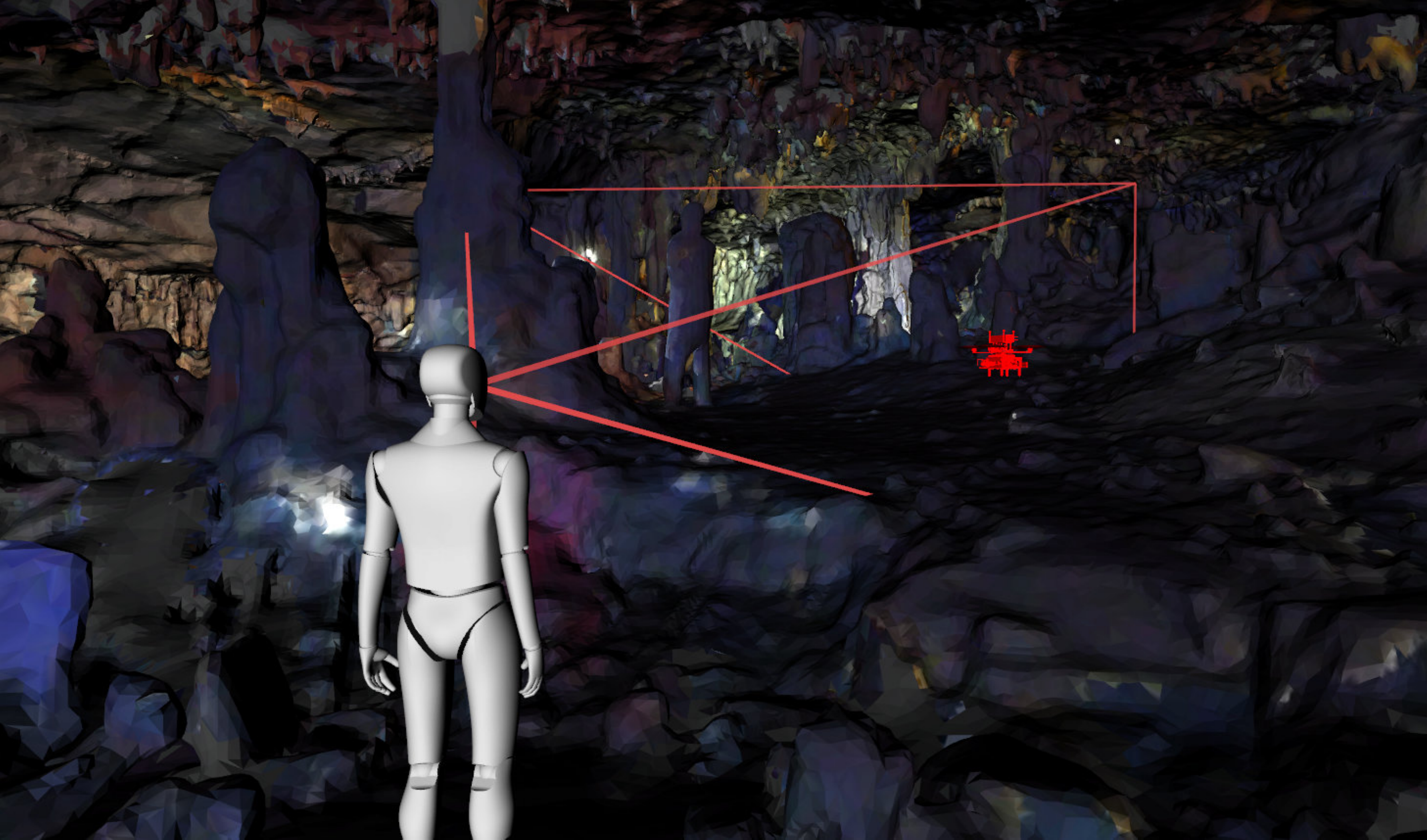}} \\
    \subfloat[\label{sfig:simroiEntropya}]{\ifthenelse{\equal{\arxivmode}{true}}{\includegraphics[]{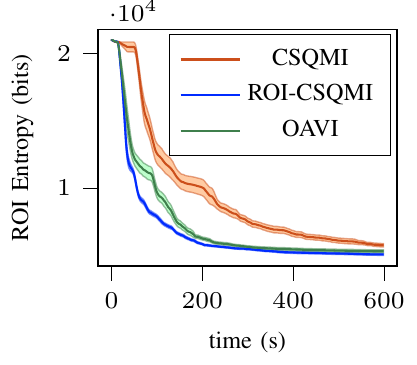}}{\input{figures/room_small_simple_wall_roiEntropy_oae_vs_csqmi_vs_csqmi_w_roi.tex}}}
    \subfloat[\label{sfig:simroiEntropyb}]{\ifthenelse{\equal{\arxivmode}{true}}{\includegraphics[]{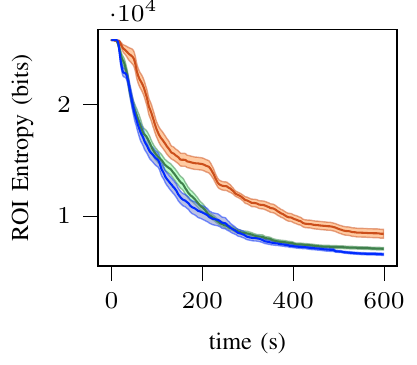}}{\input{figures/room_small_double_walls_roiEntropy_oae_vs_csqmi_vs_csqmi_w_roi.tex}}}
    \subfloat[\label{sfig:simroiEntropyc}]{\ifthenelse{\equal{\arxivmode}{true}}{\includegraphics[]{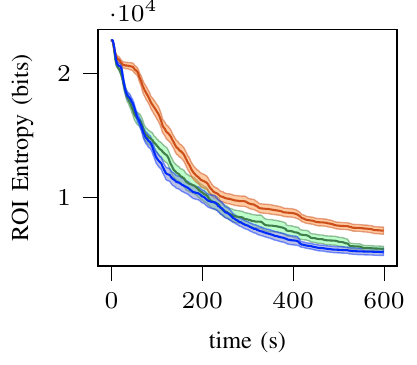}}{\input{figures/room_small_many_obstacles_roiEntropy_oae_vs_csqmi_vs_csqmi_w_roi.tex}}}
    \subfloat[\label{sfig:simroiEntropyd}]{\ifthenelse{\equal{\arxivmode}{true}}{\includegraphics[]{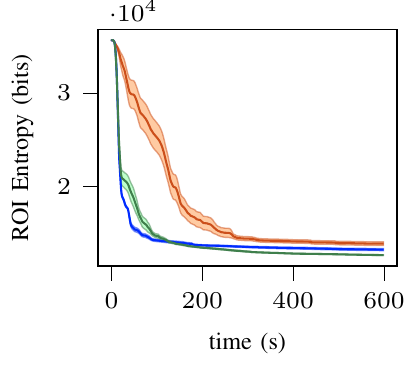}}{\input{figures/rapps_roiEntropy_oae_vs_csqmi_vs_csqmi_w_roi.tex}}} \\
    \subfloat[\label{sfig:simmapEntropya}]{\ifthenelse{\equal{\arxivmode}{true}}{\includegraphics[]{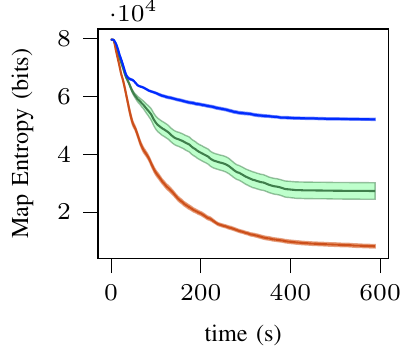}}{\input{figures/room_small_simple_wall_mapEntropy_oae_vs_csqmi_vs_csqmi_w_roi.tex}}}
    \subfloat[\label{sfig:simmapEntropyb}]{\ifthenelse{\equal{\arxivmode}{true}}{\includegraphics[]{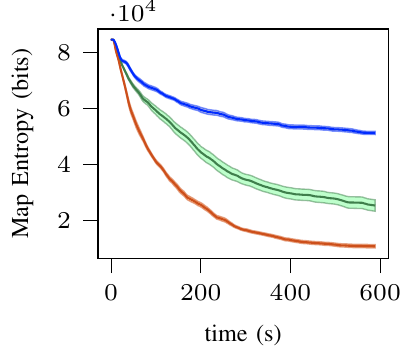}}{\input{figures/room_small_double_walls_mapEntropy_oae_vs_csqmi_vs_csqmi_w_roi.tex}}}
    \subfloat[\label{sfig:simmapEntropyc}]{\ifthenelse{\equal{\arxivmode}{true}}{\includegraphics[]{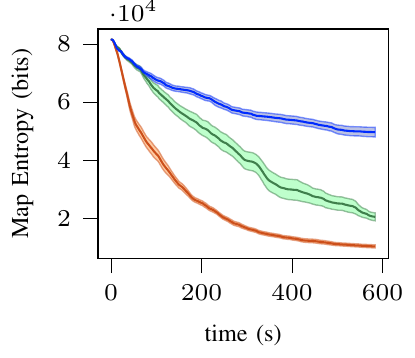}}{\input{figures/room_small_many_obstacles_mapEntropy_oae_vs_csqmi_vs_csqmi_w_roi.tex}}}
    \subfloat[\label{sfig:simmapEntropyd}]{\ifthenelse{\equal{\arxivmode}{true}}{\includegraphics[]{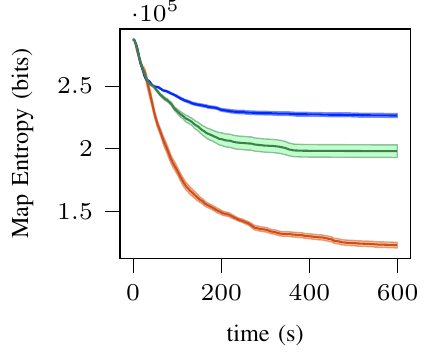}}{\input{figures/rapps_mapEntropy_oae_vs_csqmi_vs_csqmi_w_roi.tex}}}
    \caption{\label{fig:sim}
      \wennie{~\protect\subref{sfig:simenvsa}--\protect\subref{sfig:simenvsd} simulation environments,
        ~\protect\subref{sfig:simroiEntropya}--\protect\subref{sfig:simroiEntropyd} ROI entropy plotted as a function of time and
        ~\protect\subref{sfig:simmapEntropya}--\protect\subref{sfig:simmapEntropyd} map entropy plotted as a function of time for the CSQMI, ROI-CSQMI, and OAVI exploration variants.
      30 trials are run for each exploration variant and simulation
environment. Note that ROI-CSQMI and OAVI explore the human's FoV
$3\times$ faster than CSQMI while CSQMI reduces the total map uncertainty faster.
OAVI reduces the map uncertainty 56\% more than ROI-CSQMI.}}
  \end{figure*}

\section{EXPERIMENTAL DESIGN AND RESULTS}
The approach is evaluated in simulation and with real-world hardware
  experiments. The experiment begins when the human transmits the
pose of their helmet-mounted range sensor with the corresponding
pointcloud to the robot partner. Only one instance of these pose and pointcloud
pairs is transmitted for both simulation and hardware experiments.
The proposed methodology may allow for multiple pose and pointcloud
pairs, but this is left as future work.
When the robot receives data from the human, exploration begins.

The OAVI approach is compared against the ROI-CSQMI and CSQMI approaches.
Two quantitative and one qualitative measures are used to evaluate performance.
The two quantitative \wennie{evaluations} measure the entropy of the map and ROI over time.
The qualitative \wennie{evaluation} plots the evolution of the robot's trajectory over time.

\begin{table}[h]
    \begin{center}
      \begin{tabular}{|c|c|c|}
        \hline
        \textbf{Parameter} & \textbf{Simulation} & \textbf{Hardware}\\
        \hline
        Robot sensor range & \SI{5}{\metre} & \SI{2}{\metre}\\
        \hline
        Robot sensor downsampling & $2\times$ & $2\times$\\
        \hline
        Human sensor range & \SI{10}{m} & \SI{6}{m}\\
        \hline
        Human sensor downsampling & $4\times$ & $4\times$\\
        \hline
        Human FoV percentage & $40\%$ & $30\%$\\
        \hline
        Mapping frequency & \SI{10}{\hertz} & \SI{10}{\hertz}\\
        \hline
        Voxel resolution & \SI{0.3}{m} & \SI{0.2}{m}\\
        \hline
        Grid bounding box & $30\times30\times$ \SI{10}{\metre} & $4\times5\times$ \SI{2}{\metre} \\
        \hline
        Planning frequency & \SI{1}{\hertz} & \SI{1}{\hertz}\\
        \hline
        Number of motion primitives & $21$ & $15$\\
        \hline
        Max. forward velocity & \SI{0.75}{\metre\per\second} & \SI{0.40}{\metre\per\second}\\
        \hline
        Max. yaw rate & \SI{0.25}{\radian\per\second} & \SI{0.25}{\radian\per\second} \\
        \hline
        OAVI $\aroi$ & $0.10$ & $0.10$\\
        \hline
        OAVI $\apa$ & $0.15$ & $0.15$\\
        \hline
      \end{tabular}
      \caption{\label{tab:parameters}List of parameter values used in the simulation and hardware experiments.}
    \end{center}
\end{table}

\subsection{Simulation Experiments}
Simulations in four environments (\cref{fig:sim}) are conducted to
evaluate the approaches developed in this work against the baseline
approach. The simulation environments consist of a single wall,
two walls, multiple obstacles, and \wennie{the cave} environment
from \cref{fig:1} with the same human-robot placement. In
each environment the goal is for the robot to obtain views in
regions occluded to the human. These environments are selected to
highlight the merits and drawbacks of the information gain \wennie{objectives}.

In each environment, the human faces the area of interest. The human's
FoV, which is the FoV of a simulated depth camera on the human's head,
is shown as red lines in~\cref{sfig:humaninita}.  The robot is placed
at a randomly selected location within a $4\times
\SI{4}{\meter}$ box around the human's
starting position.  \wennie{After the human transmits
their pointcloud observation and pose to the robot, the robot updates
its onboard map according to~\cref{ssec:oavi}}.  \wennie{Each
exploration variant} is run for 30 trials per environment
for a total of 360 trials over all environments and
variants. Exploration is terminated after \SI{10}{\minute}
resulting in a total of \SI{60}{\hour} of simulations.

The entropy of the ROI is plotted over time for each environment
in~\cref{sfig:simroiEntropya,sfig:simroiEntropyb,sfig:simroiEntropyc,sfig:simroiEntropyd}
and the entropy of the map (including the ROI) is plotted over time
in~\cref{sfig:simmapEntropya,sfig:simmapEntropyb,sfig:simmapEntropyc,sfig:simmapEntropyd}.
In analyzing the performance
of~\cref{sfig:simroiEntropya,sfig:simroiEntropyb,sfig:simroiEntropyc,sfig:simroiEntropyd}
one can see that OAVI and ROI-CSQMI decrease the uncertainty of the ROI
approximately $3\times$ faster than CSQMI.  ROI-CSQMI slightly
outperforms OAVI because the mutual information of a view entirely
outside of the ROI is zero, which means that the robot will not select
actions outside the ROI. In contrast, \wennie{OAVI
tends to drive the robot outside the ROI after sufficient
views of the ROI have been acquired}.  When analyzing the map entropy
over time
in~\cref{sfig:simmapEntropya,sfig:simmapEntropyb,sfig:simmapEntropyc,sfig:simmapEntropyd},
one can see that the final map entropy of the OAVI approach at \SI{600}{\second} is
on average $56\%$ lower than the ROI-CSQMI approach across environments.
The baseline CSQMI approach outperforms the other approaches because
it selects views that maximize
the mutual information between the map and sensor without consideration
for the ROI.

\begin{figure*}
    \centering 
    \subfloat{\includegraphics[width=\textwidth]{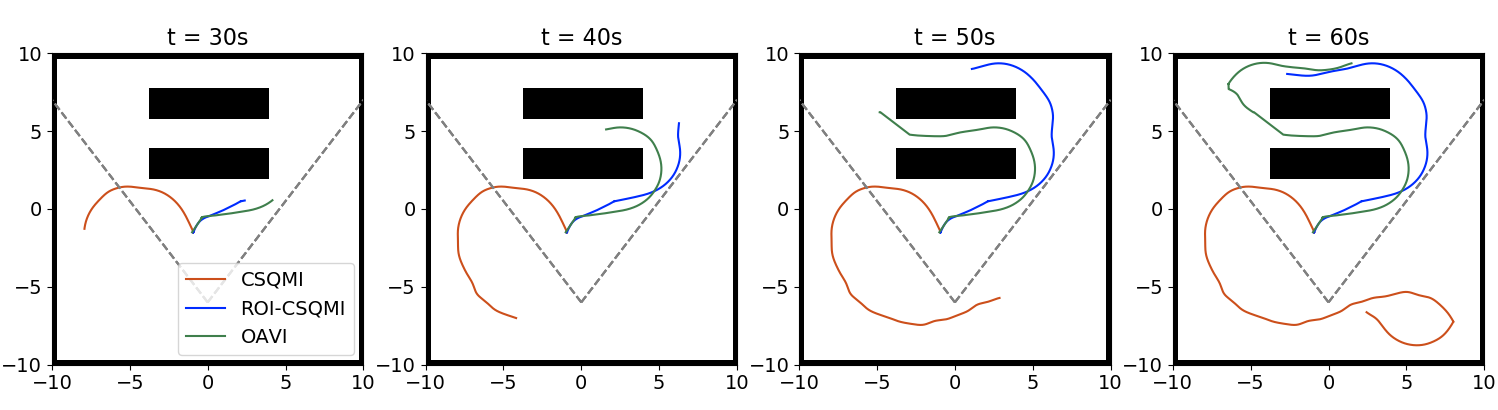}}
    \caption{\label{fig:trajectory}Top-down snapshots of the trajectory taken by the robot 
    for the three approaches in the two walls environment with the human's FoV drawn in gray dashed lines. CSQMI proceeds to explore the unknown
    regions outside of the human's FoV, while the ROI-constrained CSQMI and OAVI prioritize the ROI first.
    As opposed to ROI-CSQMI, the gradient in the OAVI approach (see \cref{sfig:oavi}) pushes the robot
    to explore the occluded region closest to the human first.}
\end{figure*} 

From these results, we arrive at the following conclusions: first, the
baseline CSQMI approach is not well suited for collaborative human-robot
exploration because it does not bias the exploration towards the
ROI; second, the ROI-CSQMI approach is ideal for a leader-follower exploration
strategy because it selects motion plans that \wennie{are} restricted within the ROI; and third,
the OAVI approach is ideal for a collaborative framework where the robot biases
views within the ROI first and then \wennie{selects} observations outside the
ROI.

\Cref{fig:trajectory} plots the top-down views of the trajectories taken
by the robot for the three approaches in the two walls
environment (shown in~\cref{sfig:simenvsb}). The evolution of the
trajectory for $t=\{30,...,60\}$\SI{}{\second} in \cref{fig:trajectory} demonstrates that
OAVI first explores the occluded region closest to the human observer and
then proceeds to the second obstacle after the robot has updated its
distance field. In comparison, ROI-CSQMI does not incorporate a measure
of the distance to obstacles so it selects actions that maximize
the mutual information between the potential observation and ROI, while CSQMI explores
areas outside of the human's ROI for the first \SI{60}{\second} because it
does not have a notion of the human's ROI.

\subsection{Hardware Experiments}
\begin{figure}[h]
  \centering
  \subfloat{\includegraphics[width=0.29\textwidth]{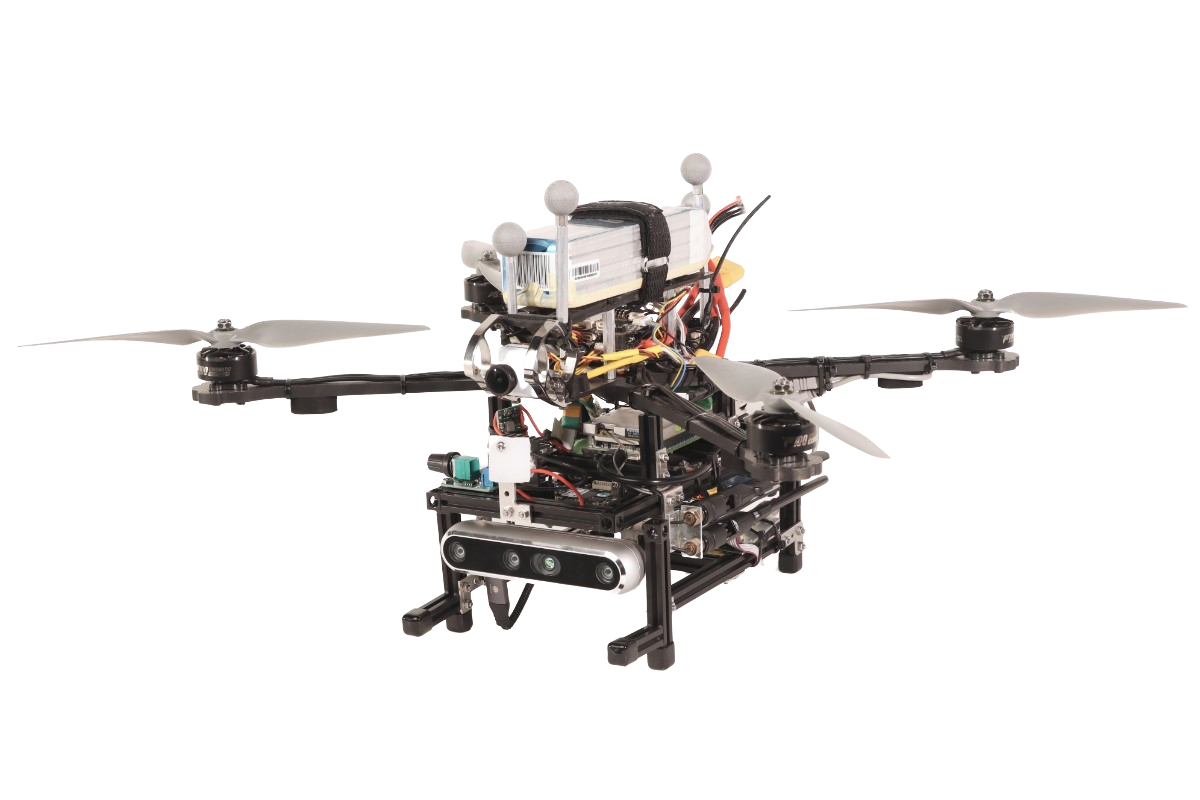}}
  \subfloat{\includegraphics[width=0.15\textwidth]{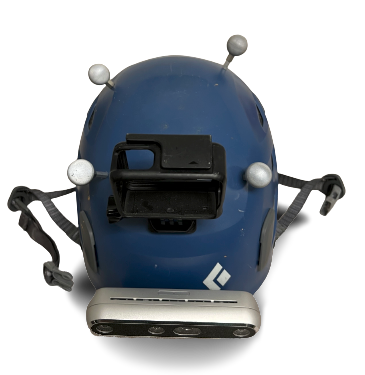}}
  \caption{\label{fig:hardware} \wennie{(Left) Aerial robot and (Right) helmet for the human partner used in the hardware experiments.}}
\end{figure}

\wennie{Experiments are run inside a motion capture arena to validate
the proposed approach against the baselines in the real-world. The
human is equipped with a helmet-mounted Intel RealSense
D455 depth camera (see~\cref{fig:hardware}).  The robot, a
\SI{2.5}{\kilogram} quadrotor,
carries a forward-facing D455 and two onboard computers: an
NVIDIA TX2 running state estimation and control, and an Intel i7 NUC 11 with 32GB of RAM,
which runs mapping, planning, and collision avoidance}.

\vspace{-0.4cm}

\begin{figure}[h!]
  \centering
  \includegraphics[width=0.8\linewidth]{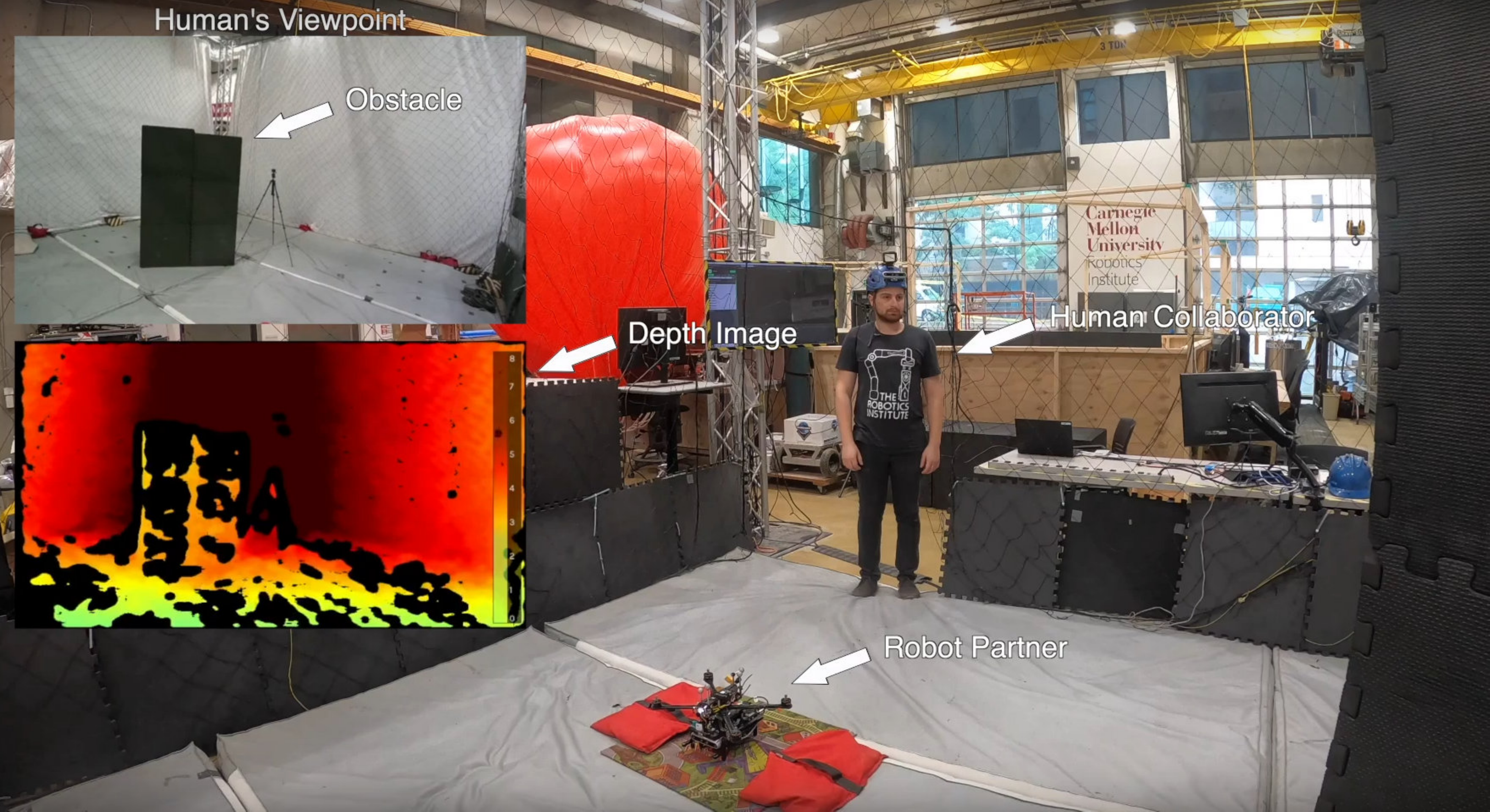}
  \caption{\label{fig:vicon}A human-robot team
    explores an environment inside a motion capture
    arena, with an obstacle in front the human requiring the robot
    to provide complementary views.}
\end{figure}

The set of parameters used
in the hardware experiments (see~\cref{fig:vicon}) are listed in \cref{tab:parameters}. Each approach is
run once starting from the same initial robot pose and a fixed helmet orientation
for a total time of \SI{2}{\minute}.

\begin{figure}[h!]
  \centering
  \subfloat[\label{sfig:hardwareroi}ROI Entropy]{\ifthenelse{\equal{\arxivmode}{true}}{\includegraphics[]{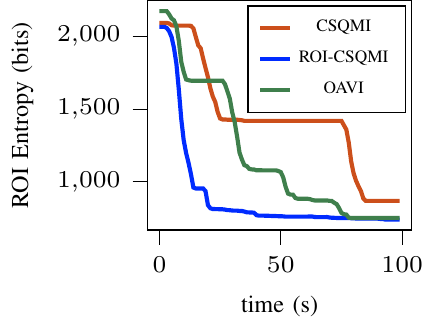}}{
\tikzsetnextfilename{hardware_roiEntropy}
\begin{tikzpicture}[font=\scriptsize]

\pgfplotsset{
        compat=1.11,
        legend image code/.code={
        \draw[mark repeat=2,mark phase=2]
        plot coordinates {
        (0cm,0cm)
        (0.15cm,0cm)        
        (0.3cm,0cm)         
        };%
        }
}

\definecolor{color0}{rgb}{0.8,0.309803921568627,0.105882352941176}
\definecolor{color1}{rgb}{0,0.168627450980392,1}
\definecolor{color2}{rgb}{0.247058823529412,0.498039215686275,0.298039215686275}

\begin{axis}[
    width=0.24\textwidth,
    height=0.22\textwidth,
    legend style={fill opacity=0.4, draw opacity=1, text opacity=1, font=\tiny},
    legend entries={CSQMI, ROI-CSQMI, OAVI},
    tick align=outside,
    tick pos=left,
    x grid style={white!69.0196078431373!black},
    xmin=-4.95, xmax=103.95,
    xtick style={color=black},
    y grid style={white!69.0196078431373!black},
    ymin=667.993167114258, ymax=2243.79815368652,
    ytick style={color=black},
    ytick = {0,1000,1500,2000},
    ylabel={ROI Entropy (bits)},
    xlabel={time (s)}
]
\addplot [very thick, color0]
table {%
0 2091.09692382812
1 2091.09692382812
2 2090.91650390625
3 2090.91650390625
4 2087.22094726562
5 2075.169921875
6 2072.58764648438
11 2072.576171875
12 2072.5419921875
13 2070.6611328125
14 2053.51611328125
15 1994.08020019531
16 1936.89294433594
17 1917.10388183594
18 1847.0244140625
19 1781.34228515625
20 1718.1904296875
21 1641.69970703125
22 1584.92810058594
23 1548.06591796875
24 1478.87976074219
25 1434.60229492188
26 1427.91674804688
27 1427.91674804688
28 1425.92822265625
32 1425.92822265625
33 1424.16198730469
34 1422.02368164062
35 1418.18530273438
51 1418.18530273438
52 1418.24230957031
67 1418.24230957031
68 1417.81665039062
72 1417.81665039062
73 1417.95446777344
74 1417.92028808594
75 1417.83764648438
77 1358.55615234375
78 1269.1201171875
79 1139.36743164062
80 1057.78149414062
81 1006.96588134766
82 969.421325683594
83 936.704895019531
84 885.280395507812
85 868.120971679688
99 868.120971679688
};
\addplot [very thick, color1]
table {%
0 2064.11645507812
2 2064.11645507812
3 2056.65063476562
4 2034.55102539062
5 1995.21081542969
6 1919.73803710938
7 1810.75183105469
8 1632.39599609375
9 1431.24157714844
10 1285.48510742188
11 1197.35327148438
12 1129.39074707031
13 1051.90686035156
14 960.209716796875
15 953.755798339844
18 953.755798339844
19 935.014831542969
20 838.959838867188
21 817.560241699219
22 812.1513671875
23 811.972961425781
24 811.688903808594
25 810.624938964844
26 810.411499023438
27 808.544494628906
28 804.872985839844
29 804.276489257812
30 802.011596679688
31 801.02978515625
32 801.023071289062
33 799.223266601562
34 798.932983398438
35 794.82080078125
36 789.839599609375
37 788.973327636719
38 788.64208984375
39 786.567138671875
40 770.109069824219
41 766.762145996094
43 766.762145996094
44 765.773559570312
45 765.773559570312
46 764.854370117188
47 764.761901855469
48 764.06103515625
49 763.082885742188
50 763.049682617188
51 761.897583007812
52 760.00244140625
53 760.135437011719
61 760.112670898438
62 761.580627441406
63 760.636962890625
64 757.62109375
65 756.508666992188
68 756.508666992188
69 757.275756835938
70 754.357788085938
71 752.529235839844
72 751.9384765625
73 751.045227050781
78 751.045227050781
79 748.987243652344
80 747.998657226562
81 747.596008300781
82 747.596008300781
83 747.337646484375
84 746.572082519531
85 746.550476074219
86 746.962707519531
89 746.94970703125
90 746.0849609375
91 744.158996582031
92 744.122375488281
93 739.757629394531
94 739.620666503906
99 739.620666503906
};
\addplot [very thick, color2]
table {%
0 2172.17065429688
2 2172.17065429688
3 2172.00048828125
4 2149.9638671875
5 2121.701171875
6 2109.02514648438
7 2068.05590820312
8 1966.25573730469
9 1823.93957519531
10 1754.89624023438
11 1702.45080566406
12 1697.34228515625
13 1694.20727539062
21 1694.19055175781
24 1694.19055175781
25 1694.1337890625
26 1693.94079589844
27 1670.63098144531
28 1621.861328125
29 1570.71704101562
30 1476.43908691406
31 1410.7783203125
32 1311.31652832031
33 1200.19372558594
34 1152.24157714844
35 1113.54162597656
36 1107.20532226562
37 1088.59204101562
38 1084.474609375
39 1084.08068847656
40 1078.22131347656
41 1078.4609375
42 1078.31103515625
43 1078.09301757812
46 1078.08154296875
48 1078.08154296875
49 1073.40209960938
50 1066.05017089844
51 1029.71362304688
52 964.455932617188
53 918.137145996094
54 911.352294921875
55 911.417541503906
56 888.152770996094
57 883.072570800781
61 883.072570800781
62 880.308044433594
63 874.350524902344
64 871.249755859375
65 871.247436523438
66 871.204895019531
67 871.408996582031
68 871.18212890625
69 871.159362792969
70 868.583801269531
71 867.484252929688
72 855.539978027344
73 846.401733398438
74 819.262573242188
75 785.800415039062
76 776.63916015625
77 774.286071777344
78 752.7431640625
79 750.967041015625
99 750.967041015625
};
\end{axis}

\end{tikzpicture}}}
  \subfloat[\label{sfig:hardwaremap}Map Entropy]{\ifthenelse{\equal{\arxivmode}{true}}{\includegraphics[]{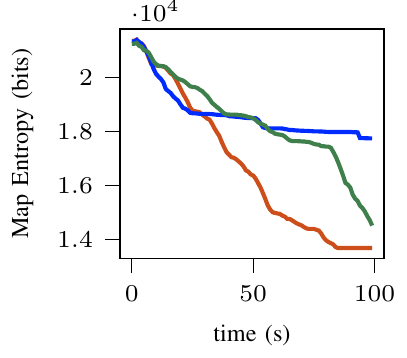}}{
\tikzsetnextfilename{hardware_mapEntropy}
\begin{tikzpicture}[font=\scriptsize]

\definecolor{color0}{rgb}{0.8,0.309803921568627,0.105882352941176}
\definecolor{color1}{rgb}{0,0.168627450980392,1}
\definecolor{color2}{rgb}{0.247058823529412,0.498039215686275,0.298039215686275}

\begin{axis}[
    width=0.24\textwidth,
    height=0.22\textwidth,
    tick align=outside,
    tick pos=left,
    x grid style={white!69.0196078431373!black},
    xmin=-4.95, xmax=103.95,
    xtick style={color=black},
    y grid style={white!69.0196078431373!black},
    ymin=13291.4651367188, ymax=21803.3961914062,
    ytick style={color=black},
    ylabel={Map Entropy (bits)},
    xlabel={time (s)}
] 
\addplot [very thick, color0]
    table {%
    0 21371.40234375
    1 21364.798828125
    2 21416.490234375
    3 21294.185546875
    4 21278.12890625
    5 21178.74609375
    6 20973.0078125
    7 20729.705078125
    8 20507.767578125
    9 20469.724609375
    10 20449.134765625
    11 20443.603515625
    12 20439.5546875
    13 20412.162109375
    14 20368.7890625
    16 20154.705078125
    17 20099.87890625
    18 19948.53515625
    19 19783.232421875
    20 19604.072265625
    21 19418.23046875
    22 19265.658203125
    23 19098.09375
    24 18896.921875
    25 18779.52734375
    26 18756.529296875
    27 18740.30859375
    28 18717.748046875
    29 18627.630859375
    30 18564.32421875
    31 18484.857421875
    32 18447.232421875
    33 18301.376953125
    34 18130.7421875
    35 17980.4375
    36 17844.451171875
    37 17622.5625
    38 17422.201171875
    39 17240.765625
    40 17148.4609375
    41 17051.888671875
    42 17029.083984375
    43 16974.26953125
    44 16895.744140625
    45 16813.953125
    46 16710.671875
    47 16562.13671875
    48 16511.49609375
    49 16411.30859375
    50 16370.2001953125
    51 16255.80859375
    53 15927.0517578125
    54 15718.2568359375
    55 15498.1845703125
    56 15259.25390625
    57 15107.1455078125
    58 15009.2314453125
    59 14979.865234375
    61 14944.7314453125
    62 14882.0029296875
    63 14846.2392578125
    64 14763.41796875
    65 14759.08984375
    66 14709.39453125
    67 14645.8984375
    68 14592.4775390625
    69 14552.671875
    70 14516.125
    71 14453.3232421875
    72 14405.3984375
    73 14389.125
    74 14388.9658203125
    75 14388.8828125
    77 14327.587890625
    78 14221.669921875
    79 14069.3759765625
    80 13966.056640625
    81 13902.7109375
    82 13859.712890625
    83 13810.8720703125
    84 13710.87109375
    85 13678.37109375
    99 13678.37109375
};
\addplot [very thick, color1]
    table {%
    0 21345.814453125
    1 21345.423828125
    2 21402.6875
    3 21328.361328125
    4 21257.71875
    5 21135.994140625
    6 20969.892578125
    7 20792.37109375
    9 20318.68359375
    10 20144.455078125
    11 20039.224609375
    12 19953.5546875
    13 19824.8125
    14 19579.197265625
    15 19504.90234375
    16 19433.56640625
    17 19314.5234375
    18 19228.349609375
    19 19156.625
    20 19016.927734375
    21 18885.921875
    22 18851.296875
    23 18794.326171875
    24 18699.162109375
    25 18681.375
    26 18678.28125
    27 18671.8671875
    28 18659.904296875
    29 18659.30859375
    30 18654.078125
    31 18651.416015625
    32 18651.41015625
    33 18649.83984375
    34 18636.896484375
    35 18626.40625
    36 18620.552734375
    38 18617.193359375
    39 18607.314453125
    40 18574.451171875
    42 18560.86328125
    43 18551.455078125
    44 18542.697265625
    45 18534.318359375
    46 18520.048828125
    47 18511.029296875
    48 18506.23046875
    49 18501.19921875
    50 18501.154296875
    51 18498.828125
    52 18439.30078125
    53 18305.380859375
    54 18163.515625
    55 18131.05078125
    56 18122.67578125
    57 18120.421875
    58 18120.611328125
    60 18120.611328125
    61 18120.216796875
    62 18116.2265625
    63 18096.74609375
    64 18080.119140625
    65 18061.6953125
    66 18059.052734375
    67 18050.708984375
    68 18041.484375
    69 18037.6328125
    70 18032.240234375
    71 18029.412109375
    72 18027.740234375
    73 18021.236328125
    74 18019.40234375
    75 18009.099609375
    76 18008.380859375
    77 18007.4765625
    78 18005.193359375
    79 17999.2734375
    80 17991.58984375
    81 17988.26171875
    82 17988.26171875
    83 17988.00390625
    84 17987.23828125
    85 17987.216796875
    86 17987.62890625
    89 17987.6171875
    90 17986.751953125
    91 17979.07421875
    92 17977.1171875
    93 17973.28125
    94 17757.767578125
    95 17758.611328125
    96 17756.521484375
    97 17753.26953125
    98 17749.376953125
    99 17749.294921875
};
\addplot [very thick, color2]
    table {%
    0 21261.064453125
    1 21260.755859375
    2 21295.373046875
    3 21185.46484375
    4 21144.654296875
    5 21020.701171875
    6 21003.78125
    7 20923.41796875
    9 20596.888671875
    10 20505.56640625
    11 20440.49609375
    12 20431.7890625
    13 20428.65234375
    14 20394.23046875
    15 20310.921875
    18 20029.314453125
    19 19965.89453125
    20 19931.791015625
    21 19902.169921875
    22 19839.828125
    23 19763.74609375
    24 19683.546875
    25 19659.314453125
    26 19650.89453125
    27 19617.947265625
    28 19554.404296875
    29 19503.259765625
    30 19407.658203125
    31 19324.400390625
    32 19208.6015625
    33 19074.013671875
    35 18916.291015625
    36 18844.154296875
    37 18758.67578125
    38 18676.05859375
    39 18643.5078125
    40 18633.529296875
    41 18631.76953125
    42 18631.619140625
    43 18630.75
    44 18615.84375
    45 18606.607421875
    46 18594.078125
    47 18571.81640625
    48 18539.275390625
    49 18527.58984375
    50 18493.29296875
    51 18433.62109375
    52 18347.416015625
    53 18292.330078125
    54 18260.1484375
    55 18219.439453125
    56 18094.85546875
    57 18014.099609375
    58 17984.435546875
    59 17921.41015625
    60 17902.767578125
    61 17887.8828125
    62 17874.81640625
    63 17821.390625
    64 17742.21484375
    65 17676.275390625
    66 17651.44921875
    67 17646.83203125
    68 17646.453125
    69 17644.453125
    70 17636.119140625
    71 17632.875
    72 17620.931640625
    73 17611.79296875
    74 17583.6015625
    75 17542.818359375
    76 17523.8515625
    77 17515.6328125
    78 17463.818359375
    79 17457.0546875
    80 17444.080078125
    81 17441.50390625
    82 17401.779296875
    83 17246.1796875
    84 17073.244140625
    85 16859.958984375
    86 16623.4765625
    87 16374.7509765625
    88 16096.501953125
    89 16032.55859375
    90 15922.466796875
    91 15658.75
    92 15512.7744140625
    93 15427.8818359375
    94 15253.6220703125
    95 15164.453125
    96 15035.353515625
    97 14863.962890625
    98 14714.759765625
    99 14513.0849609375
};
\end{axis}
    
\end{tikzpicture}}}
  \caption{\label{fig:hardwareentropy}ROI and map entropy as a function of time
    for the three approaches. The baseline CSQMI approach minimizes the total map entropy,
    while its extension ROI-CSQMI prioritizes the ROI. OAVI successfully reduces the
    uncertainty in the ROI first, followed by an exploratory behavior. A video of
    the experimental setup and
    the three exploration approaches can be found at https://youtu.be/7jgkBpVFIoE.}
\end{figure}
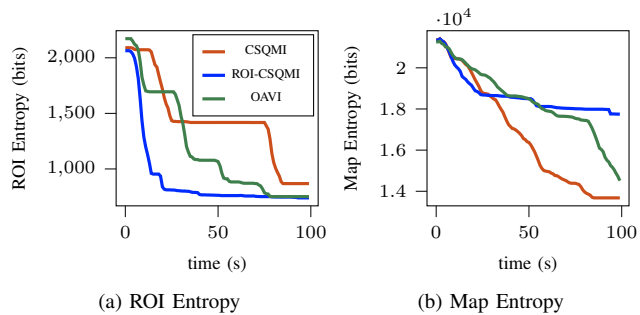

\begin{figure}
  \centering
  \includegraphics[width=\linewidth]{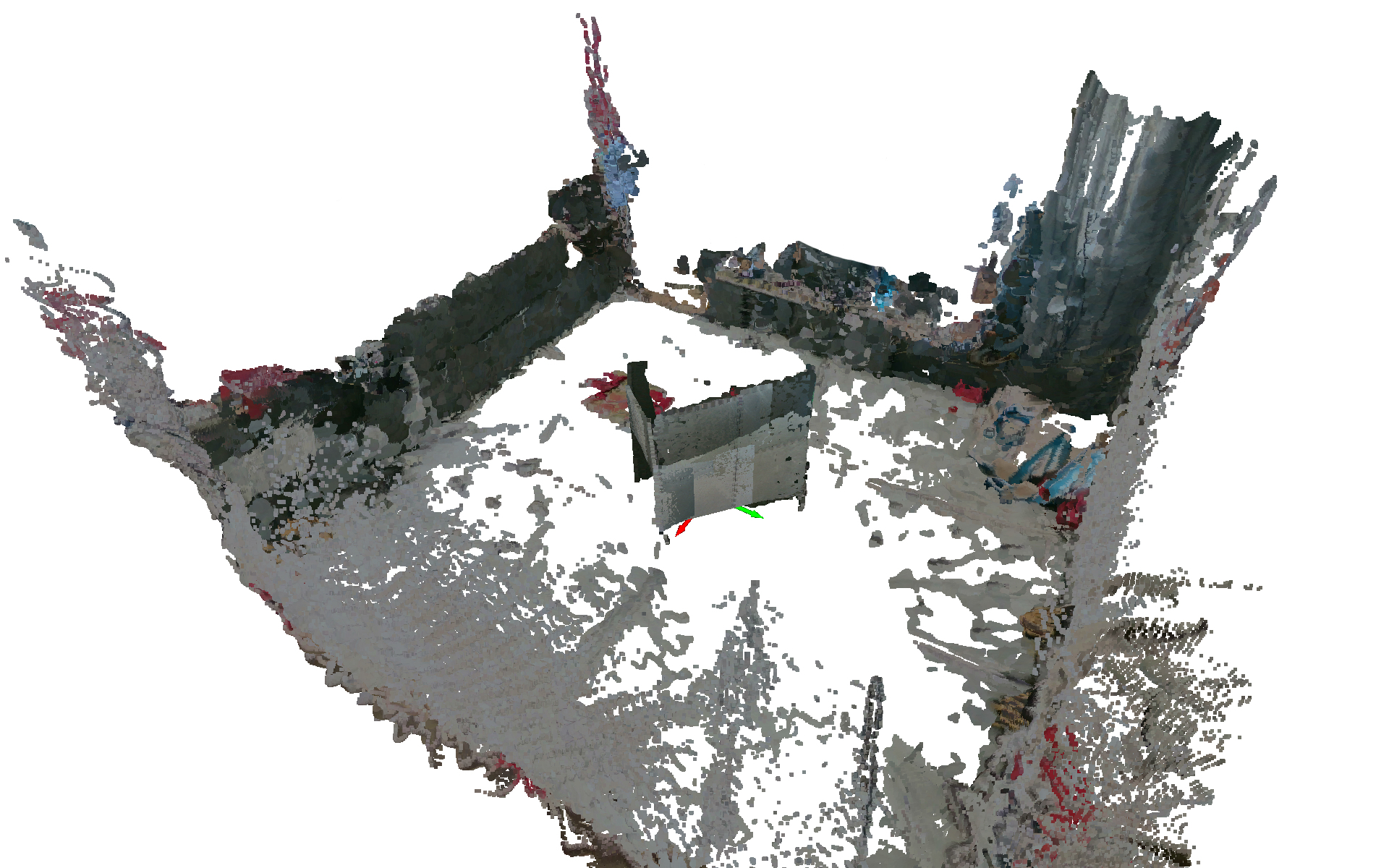}
  \caption{\label{fig:hardwarepcld}\wennie{Reconstructed point cloud map of the \SI{50}{\metre\cubed}
    environment from the OAVI hardware trial.}}
\end{figure}

The \wennie{ROI} and map entropy are plotted over time in
\cref{fig:hardwareentropy}.  \wennie{The baseline CSQMI approach
reduces the total uncertainty in the environment fastest
\cref{sfig:hardwaremap}}.  \wennie{ROI-CSQMI explores the ROI twice as
fast as OAVI, and $4\times$ as fast as CSQMI (see
\cref{sfig:hardwareroi}) but does not select actions outside of the
ROI once it reaches the ROI. This behavior yields the blue plateau
in the map at $t=\SI{50}{\second}$}. \wennie{In
contrast, OAVI explores the rest of the unknown environment as shown
in the final map (\cref{fig:hardwarepcld}), reducing the map entropy
by $75\%$ more than ROI-CSQMI.}

\vspace{0.4cm}

\begin{table}[h!]
\begin{center}
    \begin{tabular}{|c|c|}
        \hline
        \textbf{Approach} & \textbf{Planning Time}\\
        \hline
        \text{CSQMI} & $0.027 \pm 0.02$\SI{}{\second}\\
        \text{ROI-CSQMI} & $0.022 \pm 0.02$\SI{}{\second}\\
        \text{OAVI} & $0.036 \pm 0.03$\SI{}{\second}\\
        \hline
    \end{tabular}
    \caption{\label{tab:planningtimes}Planning times onboard the robot's computer \wennie{during hardware experiments} show the computational-efficiency of the proposed \wennie{approach.}}
\end{center}
\end{table}

To demonstrate computational-efficiency of the proposed approach, we record the planning
times taken by the action generation, scoring, and best primitive selection
modules onboard the robot's computer. The results reported in \cref{tab:planningtimes}
show close planning times between the three approaches, allowing our
OAVI planner to run at up \SI{15}{\hertz}.

\section{CONCLUSION}
\wennie{This paper presented a methodology for collaborative
human-robot exploration with implicit coordination.  The approach
developed in this work, OAVI, is an information-gain objective
function inspired by active reconstruction techniques.  The proposed approach
was compared against an information-theoretic exploration baseline,
CSQMI, and an extension to this baseline, ROI-CSQMI, which applies a
spatial constraint to bias actions within the human's FoV. 

Comparing these approaches in simulation and hardware
yields the following conclusions: (1) the baseline CSQMI approach is
not well-suited to the collaboration paradigm detailed in this paper because it
has no notion of the human's ROI and cannot bias motion plans to reduce
uncertainty towards the human's FoV; (2) the ROI-CSQMI approach is ideal for a leader-follower
exploration strategy because it selects motion plans that are
restricted within the ROI; and (3) the OAVI approach is ideal for the
collaborative human-robot exploration paradigm outlined in this paper
because it causes the robot to select views within the ROI first and
then explore outside the ROI when a sufficient number of views within
the ROI have been collected.}

In future work, we aim to deploy the exploration system for longer
durations and \wennie{with a moving human collaborator in outdoor,
field environments}. Further, we will relax assumptions on perfect
knowledge of human-robot poses and their relative transforms in the
world frame.


\balance

\bibliographystyle{IEEEtranN}
{
  \footnotesize
  \bibliography{refs}
  }

\end{document}